\documentclass[preprint, 3p,authoryear]{elsarticle}
\usepackage{lineno,hyperref}

\modulolinenumbers[5]

\journal{International Journal of Greenhouse Gas Control}

\usepackage{amsmath}
\usepackage{amsfonts}
\usepackage{graphicx}
\usepackage{amsthm}
\usepackage{moreverb}
\usepackage{amsfonts}
\usepackage{mathrsfs}
\usepackage{float}
\usepackage{fixltx2e}
\usepackage{gensymb}
\usepackage{xfrac}
\usepackage{lmodern}
\usepackage{color}
\usepackage{caption}
\usepackage{subcaption}
\usepackage{epstopdf}
\usepackage{listings}
\usepackage{todonotes}
\usepackage{booktabs}
\usepackage{multirow}
\usepackage{array}
\usepackage{tabularx}
\usepackage{placeins}
\usepackage{natbib}
\renewcommand{\cite}[1]{\citep{#1}}

\makeatletter
\newcommand{\thickhline}{%
    \noalign {\ifnum 0=`}\fi \hrule height 1pt
    \futurelet \reserved@a \@xhline
}
\newcolumntype{"}{@{\hskip\tabcolsep\vrule width 1pt\hskip\tabcolsep}}
\makeatother

\usepackage{array,multirow,graphicx}
\usepackage[ruled, lined, linesnumbered, commentsnumbered, longend]{algorithm2e}

\SetKwInOut{KwIn}{Input}
\SetKwInOut{KwOut}{Output}

\begin{document}

\begin{frontmatter}

\title{AI enhanced data assimilation and uncertainty quantification applied to Geological Carbon Storage}

\author[1,2]{Gabriel Serrão Seabra \corref{mycorrespondingauthor}}
\cortext[mycorrespondingauthor]{Corresponding author}
\ead{g.serraoseabra@tudelft.nl}

\author[3,4]{Nikolaj T. M{\"u}cke}
\ead{nikolaj.mucke@cwi.nl}

\author[2,5]{Vinicius Luiz Santos Silva}
\ead{v.santos-silva19@imperial.ac.uk}

\author[1,6]{Denis Voskov}
\ead{voskov@tudelft.nl}

\author[1]{Femke Vossepoel}
\ead{f.c.vossepoel@tudelft.nl}

\address[1]{Faculty of Civil Engineering and Geosciences, TU Delft, Stevinweg 1,
2628 CN Delft, Netherlands}
\address[2]{Petroleo Brasileiro S.A. (Petrobras), Rio de Janeiro, Brazil}
\address[3]{Centrum Wiskunde \& Informatica, Science Park 123, 1098 XG Amsterdam, Netherlands}
\address[4]{Mathematical Institute, Utrecht University, Utrecht, Netherlands}
\address[5]{Department of Earth Science and Engineering, Imperial College London, London, United Kingdom}
\address[6]{Department of Energy Resources Engineering, Stanford University, CA, USA}

\begin{abstract}
This study investigates the integration of machine learning (ML) and data assimilation (DA) techniques, focusing on implementing surrogate models for Geological Carbon Storage (GCS) projects while maintaining the high fidelity physical results in posterior states. Initially, we evaluate the surrogate modeling capability of two distinct machine learning models, Fourier Neural Operators (FNOs) and Transformer UNet (T-UNet), in the context of CO\textsubscript{2} injection simulations within channelized reservoirs. We introduce the Surrogate-based hybrid ESMDA (SH-ESMDA), an adaptation of the traditional Ensemble Smoother with Multiple Data Assimilation (ESMDA). This method uses FNOs and T-UNet as surrogate models and has the potential to make the standard ESMDA process at least 50\% faster or more, depending on the number of assimilation steps. Additionally, we introduce Surrogate-based Hybrid RML (SH-RML), a variational data assimilation approach that relies on the randomized maximum likelihood (RML) where both the FNO and the T-UNet enable the computation of gradients for the optimization of the objective function, and a high-fidelity model is employed for the computation of the posterior states. Our comparative analyses show that SH-RML offers a better uncertainty quantification when compared to the conventional ESMDA for the case study.
\end{abstract}

\begin{keyword}
Geological Carbon Storage (GCS), uncertainty quantification, data assimilation, machine learning. 
\end{keyword}

\end{frontmatter}



\section{Introduction} \label{section:introduction}

Geological Carbon Storage (GCS) is a vital component of Carbon Capture, Utilization, and Storage (CCUS) to mitigate greenhouse gas emissions and achieve climate targets \cite{Ringrose2019}. According to the International Energy Agency (IEA), project developers aim to bring more than 200 new capture and storage facilities into operation worldwide by 2030, with the capacity to handle over 220 Mton of CO\textsubscript{2} annually \cite{IEACCUS2022}. To put this ambitious goal into context, one of the largest CCUS projects to date, the water alternating gas (WAG) injection project in the Brazilian Pre-Salt, has injected only 20 Mton of CO\textsubscript{2} into the largest Brazilian four carbonate reservoirs (Búzios, Mero, Sapinhoá, and Tupi) over a decade \cite{CCUSBrazilROG2022}, representing less than 10\% of the IEA target. Many CO\textsubscript{2} storage sites are situated in geologically complex formations, such as fractured carbonate rocks or channelized reservoirs \cite{Burchette2012, marchAssessmentCOStorage2018}. As the number of carbon storage projects grows, it becomes crucial for companies to develop efficient forecasts and uncertainty quantification (UQ), as these studies play an important role in securing support from investors, regulators, and society during project approval and implementation.

Robust UQ and forecasts in GCS projects typically rely on several key components. These encompass robust geological models representing reservoir complexities accurately, high-fidelity reservoir simulators capturing intricate CO\textsubscript{2} injection dynamics and data assimilation (DA) techniques to combine these elements with field observations. It is important to note that effective uncertainty quantification can still be achieved even without data assimilation, depending on the specific requirements of the project and the available data.

DA techniques encompass ensemble methods (e.g., Ensemble Kalman Filters and Ensemble Smoothers), variational methods (e.g., Randomized Maximum Likelihood and 4D-Var), and fully nonlinear DA methods (e.g., Particle Filters and Markov Chain Monte Carlo) \cite{evensenDataAssimilationFundamentals2022, tarantolaInverseProblemTheory2005}. These methods leverage data, prior knowledge, and physics-based models to predict reservoir behavior under uncertain conditions. While ensemble methods are computationally efficient and flexible, variational approaches can offer better convergence but require gradient computations. Fully nonlinear methods can offer high accuracy for systems characterized by nonlinearity, such as the CO\textsubscript{2} injection. However, the computational intensity of these methods can be prohibitive due to the substantial resources and time they require. In the context of CO\textsubscript{2} injection and DA, \citet{10.1002/2014WR016264} integrated microseismic monitoring data of CO\textsubscript{2} injection with coupled flow and geomechanical models using the Ensemble Kalman Filter (EnKF), enabling the conditioning of heterogeneous rock permeability and geomechanical property distributions on microseismic data. Similarly, \citet{10.1002/2016WR020168} employed one-step ahead smoothing for joint state-parameter estimation, crucial for addressing nonlinearities in CO\textsubscript{2} storage aquifers. These studies underscore the significance of assimilating diverse data types into reservoir models, refining the understanding of subsurface properties, and optimizing CO\textsubscript{2} injection strategies, thereby contributing to the realization of effective GCS projects in mitigating climate change.

Specialized softwares are often employed to model the heterogeneities in complex reservoirs, such as channelized formations. Among these, Alluvsim is an open-source option that generates multiple geological models with key features like channel size, curvature, and shifts, using streamlines as building blocks to mimic natural deposition processes \cite{Pyrcz2009}. These detailed models reflect the heterogeneities commonly encountered in GCS projects.
For high-fidelity reservoir simulation, numerical simulators designed to handle multiphase, multicomponent flow and transport in porous media are utilized. Examples of such simulators include CMG GEM \cite{cmg2022}, SLB Eclipse \cite{eclipse2022}, and open-source options like DuMux \cite{dumux2023} and GEOSX \cite{geosx2023}. Delft Advanced Research Terra Simulator (DARTS), recently released as an open-source reservoir simulator for energy transition applications, efficiently simulates CO\textsubscript{2} injection using advanced numerical techniques like the operator-based linearization approach \cite{Lyu2023, Khait2017}. However, integrating these simulators into DA frameworks can be challenging due to high computational costs.

The choice of Alluvsim and DARTS as our primary tools stems from their proven efficacy in handling complex geological formations and fluid dynamics simulations, respectively. Alluvsim's capability to accurately simulate channelized reservoirs \cite{delottier_alluvsim}, coupled with DARTS's optimized computational efficiency \cite{Khait2017}, makes them particularly suitable for our study's objectives. This combination allows us to model the intricate interactions and variabilities within the reservoirs with a high degree of fidelity, crucial for reliable CO$_2$ injection simulations.

Recently, researchers have actively been exploring innovative strategies to merge machine learning (ML) and DA \cite{Buizza2022, silva2023generative}.  \citet{Buizza2022} provides a high-level overview of techniques for integrating DA and ML, called ``Data Learning" for improving DA in several fields. Their key focus is on approaches that leverage the strengths of ML's ability to uncover complex patterns in data and data assimilation's incorporation of physical models and dynamical constraints. Similarly, \citet{Cheng2023} explores how mixing ML and DA can make research on DA robust. The study sorts these methods into two main groups. The first group, called ``DA using ML" looks at how ML can help solve problems in data assimilation. This includes fixing errors in DA models by adding ML, using it to estimate unknown variables in DA, and defining error rates using ML methods. The study also talks about how neural networks can help in learning DA systems from start to finish. The second group, ``ML improved by DA/UQ" focuses on how DA and UQ can improve ML models. This covers topics like using Bayesian neural networks for uncertainty analysis in ML, fixing errors in simplified ML models with real-time data, and using DA to identify key equations from noisy or incomplete data. \citet{Brajard2021} recently proposed an innovative approach that combines DA and ML to infer unresolved scale parametrization in models, helping overcome limitations from sparse and noisy observational data.

In the domain of deep learning for efficient surrogate modeling, UNets have long demonstrated their efficacy, particularly in tackling subsurface problems \cite{Wen2021, Zhang2021, Pintea2021}. Originating from biomedical image segmentation, UNets excel at capturing local features through specialized convolutional layers. The architecture comprises an encoder and a decoder connected by a ``highway" system of channel concatenation, facilitating the transfer of multiscale spatial information. This has enabled outstanding predictive accuracy in diverse applications \cite{Ronneberger2015, Taccari2022}. More recently, advancements have been made by integrating UNets with transformers. \citet{Li2023} explored this method for robust medical images segmentation and \citet{AlSalmi2023} applied an attention UNet for seismic segmentation. On the other hand, Fourier Neural Operators (FNOs) have recently emerged as a promising method to build surrogate models for reservoirs submitted to CO\textsubscript{2} injection. Employing Fourier basis functions, FNOs efficiently capture multiscale interactions and offer a novel way to overcome traditional limitations in surrogate modeling \cite{Li2020, Wen2022, Witte2023}.

Recently, \citet{tang2022deep, wen2021deep, sun2019integrated, agogo2022ensemble} developed surrogate models for CCUS DA, aiming to replace physics-based numerical models. However, these models often require a substantial amount of training data from high-fidelity simulations, posing practical challenges for real-world CCUS projects with limited computational resources. While deep learning has improved surrogate model accuracy,  \citet{Dong2021} shows that it may struggle to capture subsurface complexities fully.

Hybrid models that combine physics-based and ML approaches have been explored by \citet{Tang2022, Korondi2020, deBrito2020} to mitigate the specific limitations inherent to both physics-based and ML models, with the goal of forging a more balanced and comprehensive tool. ML models, while proficient at identifying patterns and correlations within large datasets, may lack the capability to infer the underlying physical processes governing the system. This limitation can lead to potential inaccuracies in predictions under unseen conditions or parameter ranges, especially when dealing with the intricate geological variations and non-linear fluid dynamics inherent in subsurface environments. On the other hand, physics-based models offer reliable insights into these underlying processes but may struggle with computational efficiency. However, significant challenges arise when incorporating ML surrogate models in DA due to the different parameterizations employed by ML surrogates and physics-based, high-fidelity simulators. This misalignment can impede the integration, limiting the applicability and efficacy of the resulting UQ in real-world GCS projects.

In our methodology, we initially employ a standard ESMDA approach utilizing DARTS for high-fidelity simulations of channelized reservoirs built with Alluvsim. This scenario poses a significant challenge for DA due to its highly nonlinear nature and the non-Gaussian distribution of parameters. To improve upon the standard ESMDA methodology, we evaluate two ML surrogate models for comparison: one rooted in the Fourier Neural Operators and the other adopting a Transformer UNet (T-UNet) architecture, which to our knowledge is the first application of these techniques to GCS subsurface problems. Our observations reveal that FNOs show a slight advantage over T-UNet, particularly for small datasets. Subsequently, we develop two hybrid techniques to integrate DA with these ML surrogates. The first, termed Surrogate-based hybrid ESMDA (SH-ESMDA), incorporates the ML surrogates models, expediting the ESMDA process by around 50\% or more, and thereby facilitating quicker uncertainty evaluations. For the second technique, known as  Surrogate-based Hybrid RML (SH-RML), we use the ML surrogates models specifically for gradient calculations within a variational framework and compute the posterior curves with high-fidelity physics simulator DARTS. The SH-RML achieves better history matching than ESMDA and SH-ESMDA.

In summary, our contributions are as follows:

\begin{itemize}
    \item We train and test two different types of novel ML surrogates in a channelized reservoir setting for CO\textsubscript{2} storage, a FNO and a T-UNet.
    \item We introduce two novel hybrid methods, SH-ESMDA and SH-RML, that incorporate ML into both ensemble and variational DA techniques. The first, significantly expedites the DA process and the second allows one to perform variational DA.
    \item Both proposed methods ensure that posterior high-fidelity physics solution is respected.
    \item The proposed methods are versatile and can be adapted to various physical systems beyond CO\textsubscript{2} sequestration.
\end{itemize}

The paper is organized as follows: Section \ref{section:reservoir_modeling} discusses the creation of geological models, followed by CO\textsubscript{2} injection simulations with DARTS. In Section \ref{section:neural_networks}, we delve into the ML models used to build the surrogate models. Section \ref{section:data_assimilation} and Section \ref{section:hybrib_DA} describes the DA methods discussed in the paper, ESMDA, RML and the hybrid methods SH-ESMDA and SH-RML. Finally, Section \ref{section:results} presents the results and implications of these methods for enhancing DA in CO\textsubscript{2} storage projects.

\section{Overview of the Reservoir Model for CO\textsubscript{2} Injection} 
\label{section:reservoir_modeling}
\subsection{Geological Modeling Using Alluvsim}

We use Alluvsim, a specialized algorithm to simulate channelized reservoirs \cite{Pyrcz2009, delottier_alluvsim}. This tool allows us to manipulate different geological variables that impact channel features. Following the guidelines by \citet{Pyrcz2009}, we create multiple geological models by altering essential parameters within certain limits.

We consider as variables the likelihood of channel shifts, known as avulsion probability, and vertical sediment build-up, or aggradation levels, within statistically defined ranges. Parameters such as channel orientation, thickness, and geometric aspects like the width-to-thickness ratio are also modeled using various distributions to mimic natural variability. We similarly vary levee width to represent lateral sediment deposition and channel sinuosity to capture meandering behavior. Properties are distributed across different facies, to accurately represent variations in rock quality. This allows a comprehensive evaluation of the reservoir's attributes while accounting for uncertainty.

\begin{figure}[!h]
\centering
\includegraphics[width=0.7\linewidth]{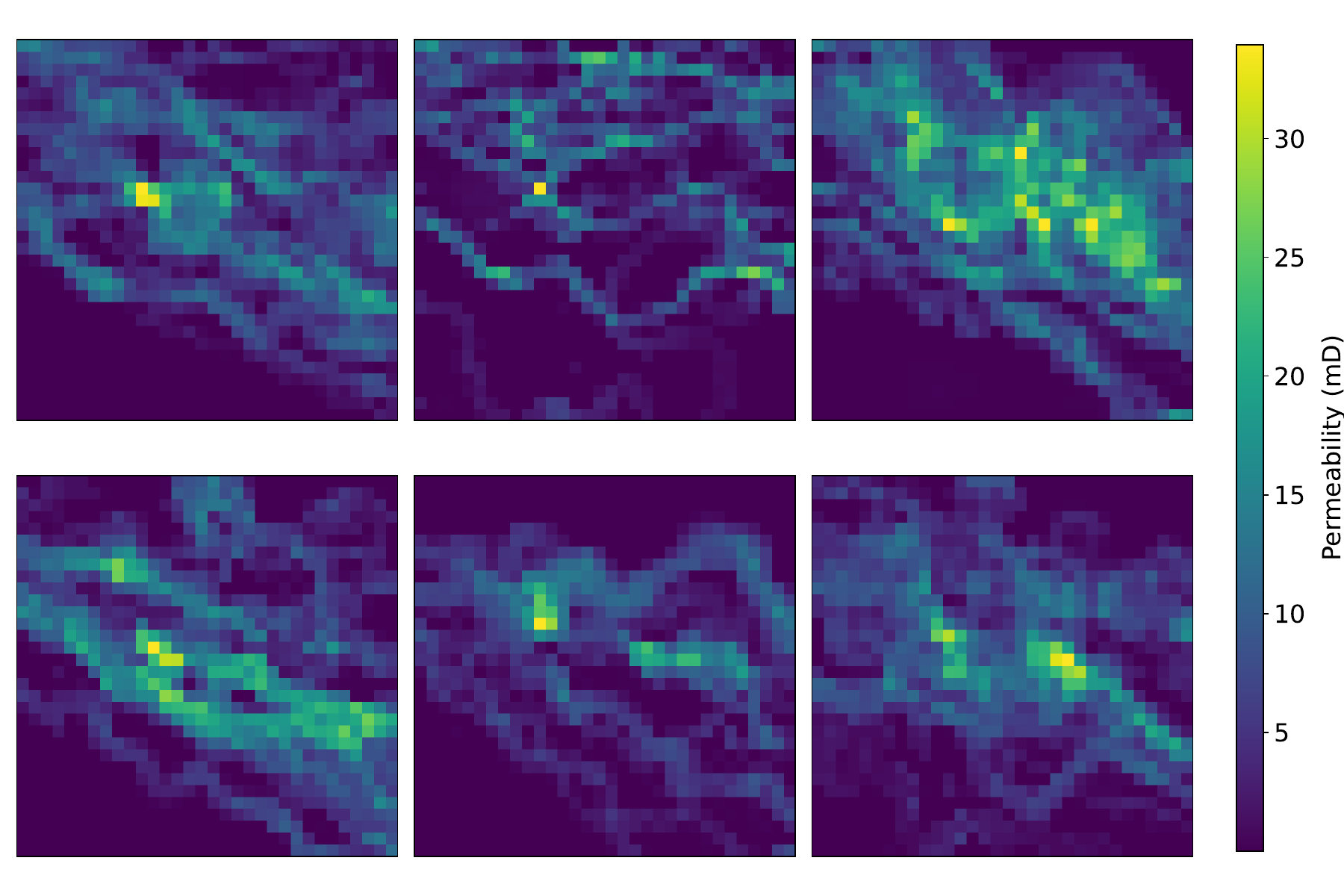}
\caption{Permeability maps of six models from the dataset.}
\label{fig:geological_models}
\end{figure}

By randomly selecting values for the aforementioned parameters, we produce multiple realizations that capture the variability in the properties of channelized reservoirs. Figure~\ref{fig:geological_models} showcases the permeability distributions of six randomly sampled models from this dataset. Each model in the dataset has a grid dimension of \(32 \times 32 \times 1\) with a spatial discretization of \(192 \times 192 \times 10\) m. This grid resolution was carefully chosen, considering computational efficiency and memory requirements.

\begin{figure}[!h]
\centering
\includegraphics[width=0.7\linewidth]{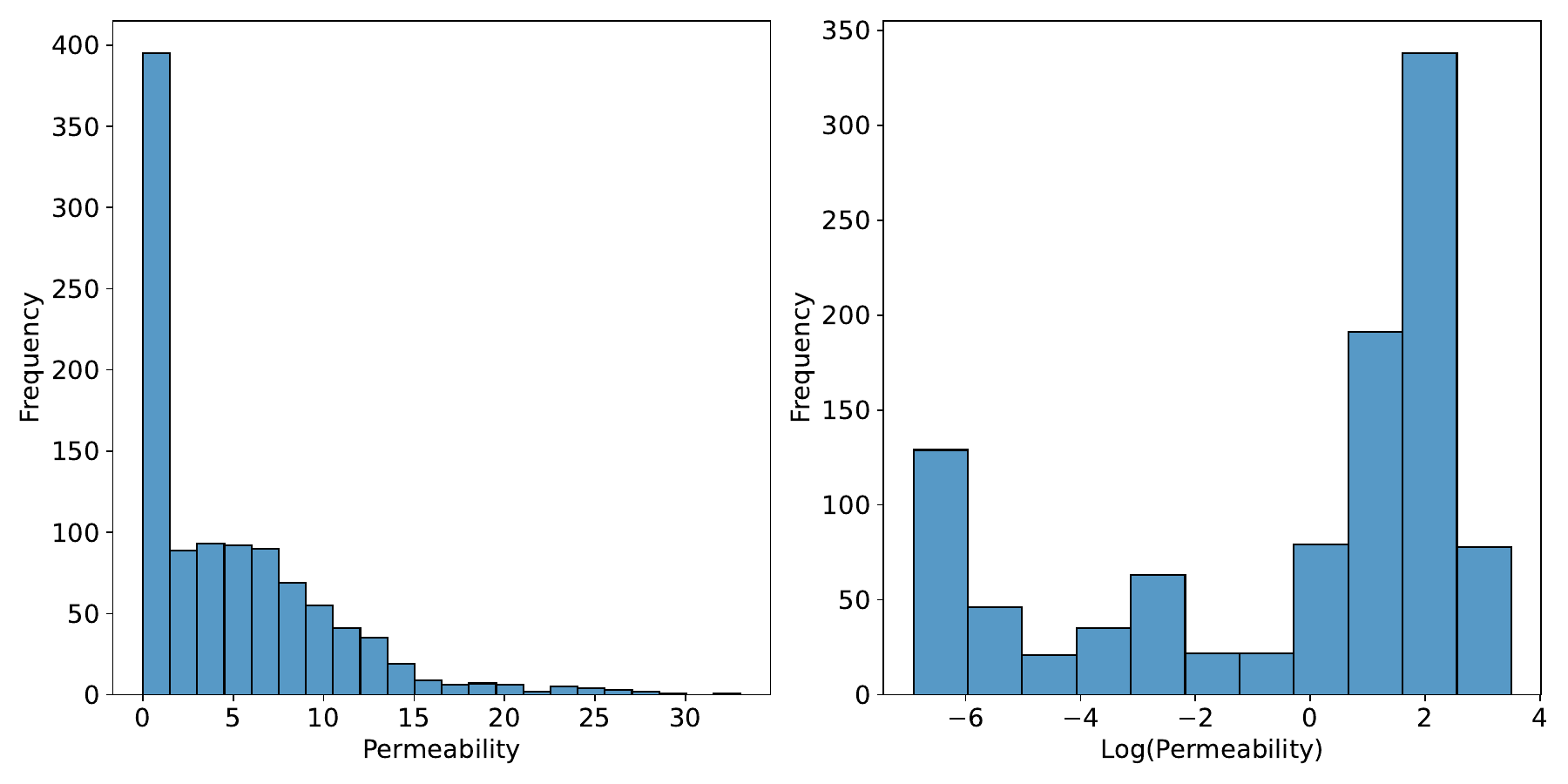}
\caption{ Histogram of permeability (left) and log of permeability (right) for one sample.}
\label{fig:permeability_histogram}
\end{figure}

The histograms in Figure~\ref{fig:permeability_histogram} clarify the permeability distribution of a model sample from our dataset. Note that the distributions are not Gaussian. These diverse geological models define the permeability distribution for subsequent CO\textsubscript{2} simulations which will, in turn, be used for the training of the ML methods. 

\subsection{CO\textsubscript{2} Injection Simulation Using DARTS}

Delft Advanced Research Terra Simulator (DARTS) is used to perform reservoir simulations of CO\textsubscript{2} injection into channelized reservoirs. DARTS is engineered for handling complex flow through porous media and is optimized for computational speed via techniques such as operator-based linearization (OBL) rendering a high-fidelity, physics-based representation \cite{Khait2017, DARTSpour2023nonlinear, DARTSchen2020optimization}. The simulator employs the Peng-Robinson equation of state to compute key fluid properties like density, viscosity, and enthalpy \cite{Lyu2023}. In DARTS, the geological model is discredited as a computational grid. Within this grid, one well is placed in the center of the reservoir for CO$_2$ injection. The model simulates the injected CO$_2$ plume and conducts equilibrium flash calculations to determine phase partitioning. 

\begin{figure}[!h]
\centering
\includegraphics[width=0.7\linewidth]{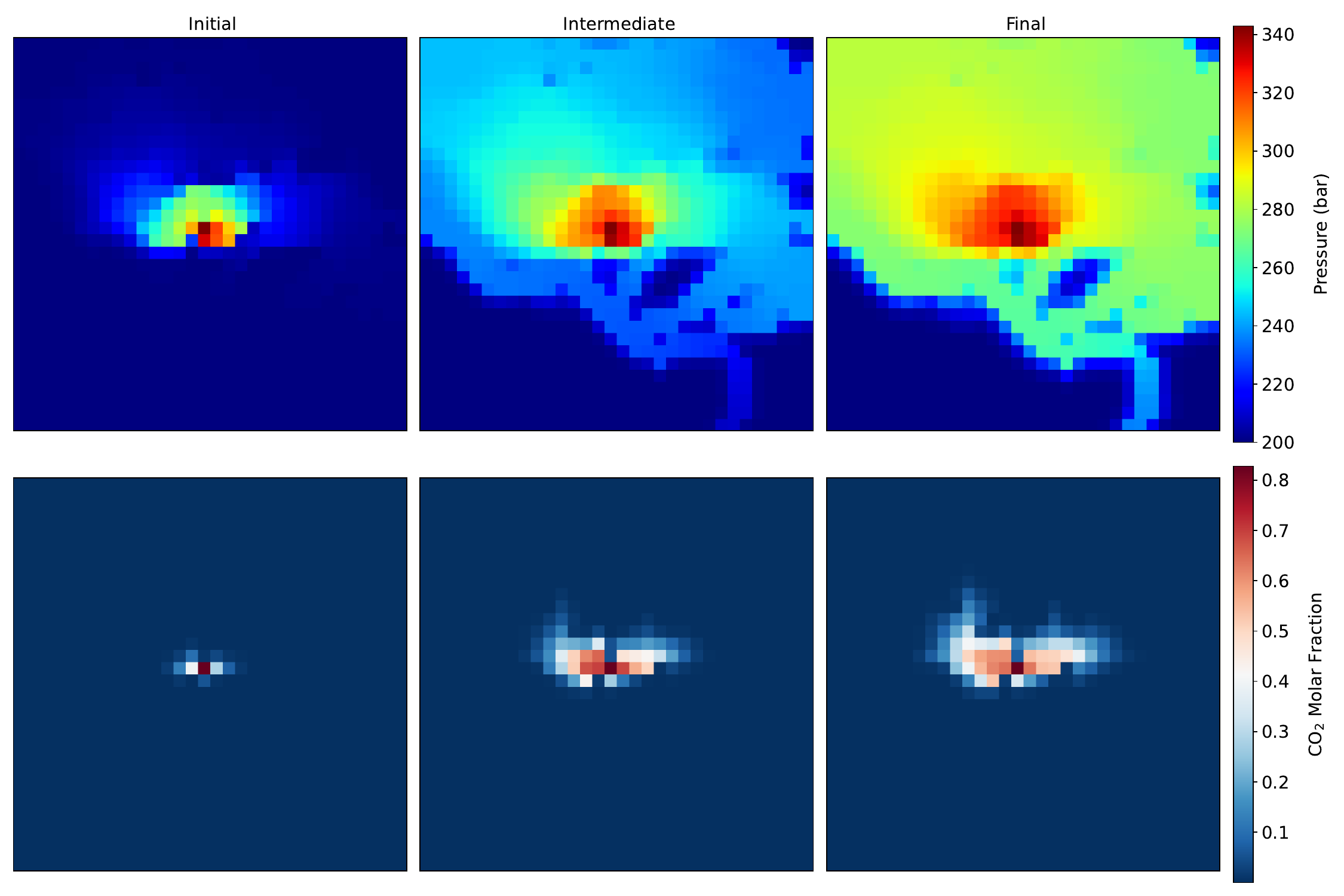}
\caption{Simulation results using DARTS. Top row: Pressure distributions; Bottom row: CO\textsubscript{2} molar fractions.}
\label{fig:darts_example}
\end{figure}

Our simulation includes CO\textsubscript{2}, CH\(_4\), and H\(_2\)O as components, with initial conditions featuring a uniform gas saturation of 20\% and a composition of 2\% CO\textsubscript{2} and 98\% CH\(_4\). CO\textsubscript{2} is injected via one well in the center of the reservoir at a time-varying prescribed gas rate. In our simulations, the time frame comprises 61 time steps, each lasting 30 days. Although this is a shorter window than what is typically encountered in real-world CCUS projects, this duration is sufficient to induce overpressure in the reservoir, which is an aspect we aim to investigate as it can impact the further CO\textsubscript{2} distribution in the reservoir. Figure \ref{fig:darts_example} illustrates our simulation results. The top row shows pressure distributions, and the bottom row displays CO\textsubscript{2} molar fractions, captured at the initial, intermediate, and final time steps.

A key observation is the differential progression of the pressure and CO\textsubscript{2} fronts. Pressure changes, governed by diffusion phenomena propagate more rapidly through the reservoir. In contrast, the CO\textsubscript{2} front advances more slowly, influenced by complex transport mechanisms. This differential movement highlights the potential value of monitoring pressure as an early warning system for subsurface changes, even before significant CO\textsubscript{2} migration occurs. Because of this, we choose pressure as a critical variable for monitoring in our subsequent DA studies.

By applying DARTS to the Alluvsim realizations, we generate a comprehensive suite of high-fidelity simulations that form the basis for our subsequent ML model training and DA study. It's important to note that the combination of geological complexities modeled by Alluvsim and the fluid dynamics simulated by DARTS creates a highly nonlinear problem. Coupled with the non-Gaussian distribution parameters, as evident from Figure~\ref{fig:permeability_histogram}, this poses a substantial challenge for traditional DA methods. This complexity further underscores the need for advanced approaches, including ML-based techniques, to accurately perform DA. 

\section{Neural Network as Surrogate Forward Models}
\label{section:neural_networks}
A surrogate model is a model that replaces the high-fidelity model for the simulation of fluid behavior in a reservoir. The surrogate model should be computationally cheap to evaluate without significantly sacrificing accuracy. To achieve this, we construct a surrogate model based on the output of the high-fidelity model. This is done in two stages: an \textit{offline stage}, in which the surrogate model is trained, and an \textit{online stage}, where the surrogate model is used in place of the high-fidelity model. Conventionally, surrogate models achieve speed-ups by lowering the dimension of the state and corresponding equations to be solved, such as in proper orthogonal decomposition \cite{quarteroni2015reduced, hesthaven2016certified}. However, due to large Kolmogorov $N$-widths for highly nonlinear and/or hyperbolic problems \cite{ohlberger2015reduced}, such linear approaches have lately been replaced with equation-free methods. Here, one replaces the equations with a function that directly gives the output of interest. For this to work, one must typically use highly complicated functions to make up for the lack of physical knowledge in the online stage. The offline stage is typically significantly more expensive and requires more training data. Neural networks are immensely popular in this approach due to their capability of approximating highly nonlinear functions \cite{mucke2021reduced, Li2020, geneva2022transformers}. 

The general setup for the offline stage is to first generate high-fidelity solutions and then train the neural networks on these solutions. In this paper, we specifically make use of the neural network architectures T-UNet and FNOs. 

\subsection{Forward Model}
The forward model, that is, the simulator used to describe the behavior of fluid in a reservoir, maps input parameters to a state trajectory. In our case, it maps permeability and porosity to the state trajectory of pressure and fluid flow. Injection rate is the control that determines the behavior of the fluid. Let $\boldsymbol{K}\in\mathbb{R}^{N_x\times N_y}$ be the permeability, $\boldsymbol{\phi}\in\mathbb{R}^{N_x\times N_y}$ the porosity, $\boldsymbol{q}\in \mathbb{R}^{N_t}$ the injection rate, and $\boldsymbol{m}=(K, \phi, q)$. We then define the space of parameters $\boldsymbol{m} \in M$. The state trajectories consist of pressure, $\boldsymbol{p}\in\mathbb{R}^{N_x\times N_y\times N_t}$, and $CO_2$ molar fraction, $\boldsymbol{f}\in\mathbb{R}^{N_x\times N_y\times N_t}$. We furthermore define the state trajectory space, $\boldsymbol{d}\in V$ and find that $\boldsymbol{d}=(\boldsymbol{p},\boldsymbol{f})$. With this, the forward map, $G$, is defined by:
\begin{align}
    G: M \rightarrow V, \quad \boldsymbol{m} \mapsto \boldsymbol{d}.
\end{align}
The surrogate forward model, $\hat{G}$, approximates $G$:
\begin{align}
    \hat{G}: M \rightarrow V, \quad \boldsymbol{m} \mapsto \boldsymbol{d}, \quad \hat{G}(\boldsymbol{m})\approx G(\boldsymbol{m}).
\end{align}
While $G$ maps $\boldsymbol{m}$ to $\boldsymbol{d}$ implicitly by solving a set of PDEs, $\hat{G}$ directly maps $\boldsymbol{m}$ to $\boldsymbol{d}$. $\hat{G}$ is parametrized by a family of neural networks. As $\hat{G}$ is a neural network, it consists of a set of weights, $\boldsymbol{\theta}$. The weights are fitted in the offline stage. 

The training of $\hat{G}$ is performed in the offline stage by first generating $N_s$ training samples by using the high-fidelity forward model:
\begin{align}
    M_{\text{train}} = \left\{ \boldsymbol{m}_i   \right\}_{i=1}^{N_s}, \quad 
    V_{\text{train}} = \left\{ G(\boldsymbol{m}_i)   \right\}_{i=1}^{N_s} = \left\{ \boldsymbol{d}_i   \right\}_{i=1}^{N_s}, \quad S_{\text{train}} = (M_{\text{train}}, V_{\text{train}}).
\end{align}
$\hat{G}$ is then trained by minimizing a loss function with respect to the weights of $\hat{G}$, $\theta$:
\begin{align}
    L(\hat{G}, S_{\text{train}}) =  \frac{1}{N_s}\sum_{i=1}^{N_s} l( \hat{G}(\boldsymbol{m}_i), \boldsymbol{d}_i ) + \lambda ||\boldsymbol{\theta}||_2^2, \quad (\boldsymbol{m}_i, \boldsymbol{d}_i) \in S_{\text{train}},
\end{align}
where $l$ is some loss, typically the $l^p$ or $L^p$ norm of the residual, $\lambda$ is a hyperparameter, and $||\cdot||_2^2$ is the squared $l^2$ norm and serves as a regularization term. The minimization of $L$ is performed by stochastic gradient descent (SGD) with respect to $\boldsymbol{\theta}$. The specific SGD algorithm is often chosen to be the Adam optimizer \cite{kingma2014adam} or other variations thereof. 

Below, we will present the particular neural network architectures we use in this paper. 

\subsection{Transformer UNet}
The UNet architecture was introduced in \citet{Ronneberger2015}. The idea is to reduce the dimension of the input data down to a bottleneck via a series of convolutional layers and then increase the dimension back to the original shape via upscaling convolutional layers. The bottleneck layers serves as a low-dimensional representation of the data that is rich in feature information. In the upscaling part of the network, the intermediate layers from the downscaling part are concatenated to the convolutions to provide context.  

Originally developed and primarily utilized in the medical imaging field, the T-UNet architecture is being adapted in this work for GCS projects. To the best of our knowledge, this constitutes the first endeavor to apply the T-UNet architecture in the realm of GCS. 

The choice of the T-UNet architecture, a variant of the well-established UNet model, is grounded in its substantial application and success in various fields, particularly in subsurface applications \cite{Ronneberger2015, AlSalmi2023}.

The UNet type of architecture allows one to add information to the predictions on various spatial levels. Specifically, we utilize this to inform the forward model with the injection rate in the bottleneck layers. By adding this information in the bottleneck layers, we effectively affect the rich feature encodings with additional information in an efficient manner. This makes the computations cheaper and makes it easier for the network to learn the relations between the input parameters and the output. 

In the proposed architecture, we input the spatially distributed parameters, porosity and permeability, as a two-channel ``image". Then we copy that $N_t$ times and concatenate a channel consisting only of the time. This way we can compute the bottleneck encoding of the space for each time step as a batch consisting of 3D tensors (channels, height, width) rather than a single 4D tensor (channels, time steps, height, width), which enables us to use 2D convolutional layers instead of 3D convolutional layers. Since 3D convolutional layers are significantly more memory and compute-heavy, this gives us significantly more efficiency.

For the conditioning of the bottleneck layer, we use cross-attention in the shape of the transformer architecture \cite{vaswani2017attention}. The transformer has been shown to provide state-of-the-art performance on multiple types of data and seems to be the superior choice for multi-modal data \cite{rombach2022high, xu2023multimodal}. Unfortunately, the attention mechanism is very compute and memory intensive and scales poorly with the data dimension. By utilizing the transformer in the bottleneck layers, however, this problem is circumvented. For a description of the transformer neural network, see \cite{vaswani2017attention}.

\begin{figure}
    \centering
    \includegraphics[width=1.0\linewidth]{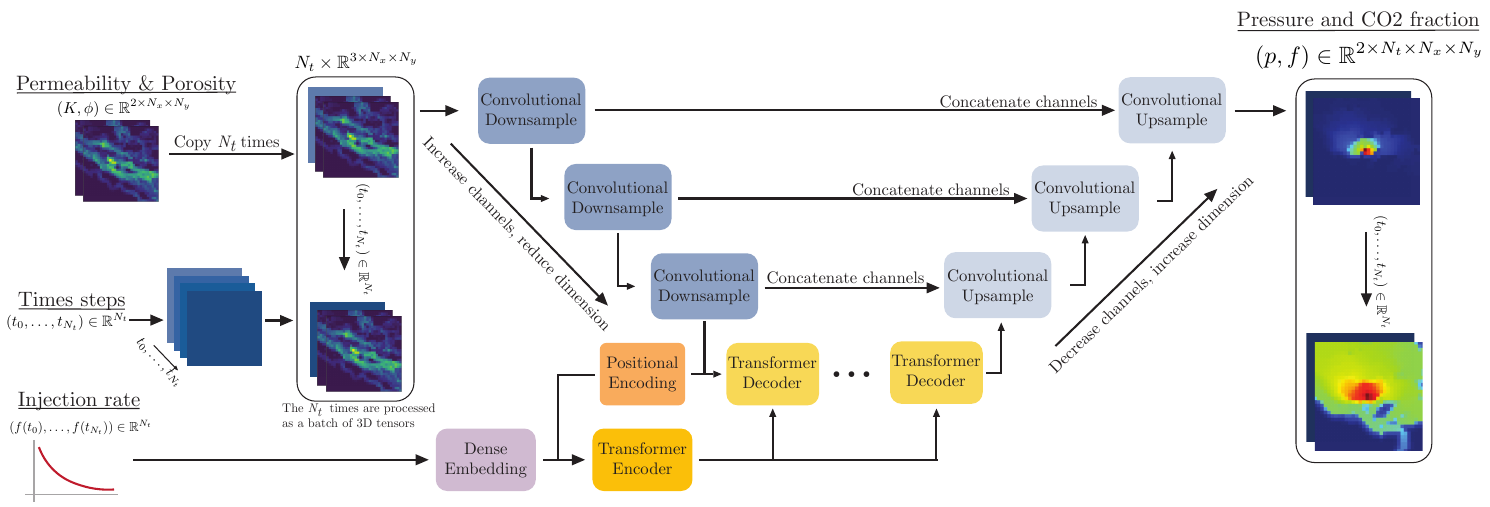}
    \caption{T-UNet architecture. }
    \label{fig:transformer}
\end{figure}

As mentioned, we utilize transformers to condition the forward model on the injection rate. We do this by first embedding the injection rate time series through dense layers, such that the dimensions match the encoded spatial dimensions. Then, we employ positional encoding which receives information from a dense embedding that originates from the gas injection rates. The encoded positions are then passed to the transformer decoder layers, effectively providing a richer context for each time step. This ensures that the transformer is not only aware of the feature information but also the sequence in which they occur. The embedding time series is passed through transformer encoder layers after which it is combined with the spatial data through transformer decoders. For a visualization of the full T-UNet, see Figure~\ref{fig:transformer}

\subsection{Fourier Neural Operators}
\begin{figure}
    \centering    \includegraphics[width=1.0\linewidth]{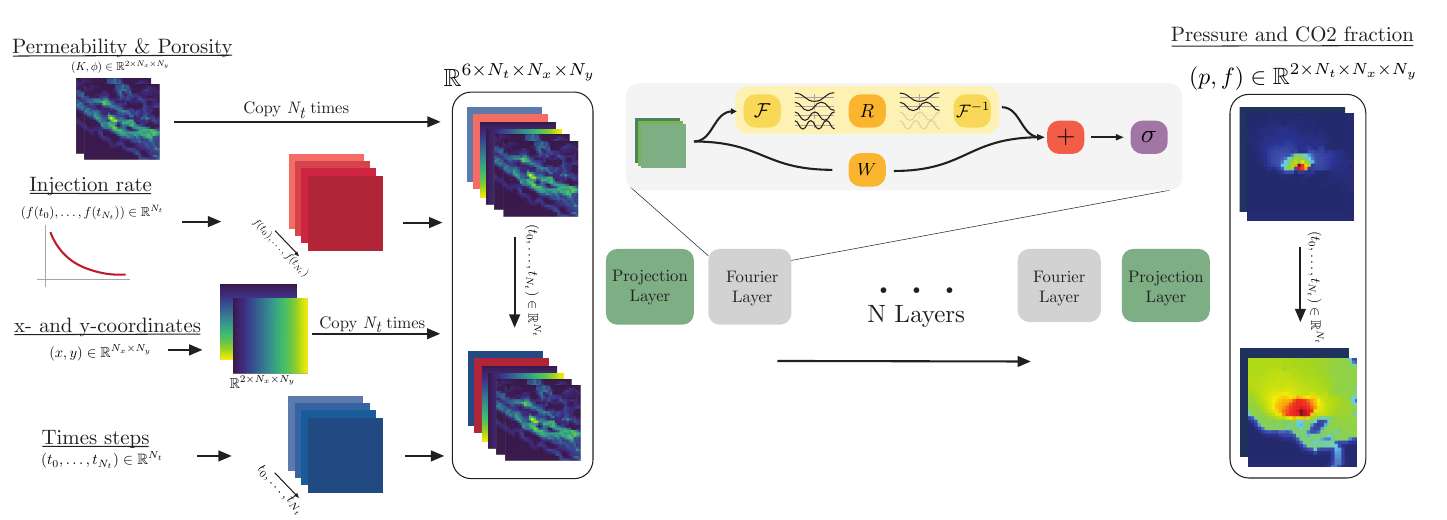}
    \caption{FNO architecture.}
    \label{fig:FNO}
\end{figure}

FNOs were introduced in \cite{Li2020} for various parametric PDE problems. In contrast to conventional neural networks, FNOs learn operators between function spaces instead of Euclidean spaces. This makes FNOs resolution invariant. The general idea is to make use of the Fourier transform, followed by a series of operations in Fourier space, after which the data is transformed back to physical space. Hence, a single Fourier layer is given by:
\begin{align}
    a^{n+1}(x) = \sigma\left( W a^{n}(x) + \mathcal{F}^{-1}(R \cdot (\mathcal{F}a^{n}))(x)  \right),
\end{align}
where $a^{n}(x)$ is the output of the $n$th layer, $W$ and $R$ are affine transformations consisting of trainable weights, $\mathcal{F}$ is the Fourier transform, and $\sigma$ is an activation function. Before applying $R$ the number of modes is truncated to a pre-defined number of modes, $k$. The Fourier transform is in practice approximated by a discrete Fourier transform. $W$ is typically a standard convolutional layer with a kernel of size one. While truncating the number of Fourier modes removes high-frequency information, the convolutional layer, $W$, compensates for that. For visualization of the FNO layer, see Figure~\ref{fig:FNO}

The FNO layers are preceded by a projection layer, that maps the number of input channels to the desired number of hidden channels. Similarly, the FNO layers are superseded by another projection layer that maps the number of hidden channels to the number of output channels. 

For our specific application, we use the FNO to map parameters, $\boldsymbol{m}$, to corresponding state trajectories, $\boldsymbol{d}$. To capture the 3D structure of the data, we use 3D Fourier transform and 3D convolutions. The injection rate is encoded to have a 3D structure. The rate at each time step is copied onto all discrete spatial points:
\begin{align}
    \boldsymbol{q}_{\text{enc}} = [\boldsymbol{q}_{0} \boldsymbol{1}_{N_x\times N_y}, \ldots, \boldsymbol{q}_{N_t} \boldsymbol{1}_{N_x\times N_y}],
\end{align}
where $\boldsymbol{1}_{N_x\times N_y} \in \mathbb{R}^{N_x\times N_y}$ is a matrix consisting of ones. The subscript $()_{\text{enc}}$ signifies that the quantities are encoded to fit the 3D tensor format. Similarly, the spatial points coordinates, $(x, y)$ are encoded and copied along the temporal dimension. The time steps are treated in the same way as the injection rate. Lastly, the porosity and permeability are also copied along the temporal dimension. Hence, the input to the FNO is:
\begin{align}
    (K, \phi, q, x, y, t)_{\text{enc}} = (K_{\text{enc}}, \phi_{\text{enc}}, q_{\text{enc}}, x_{\text{enc}}, y_{\text{enc}}, t_{\text{enc}}) \in \mathbb{R}^{N_c\times N_x\times N_y\times N_t},
\end{align}
where $N_c$ is the number of channels. In our case, $N_c=6$ -- permeability, porosity, injection rate, $x$, $y$, and time step. 

For the training of the FNO, we use the squared $L^2$-norm. This is an unusual choice for neural networks but a very common metric for PDEs. As neural networks typically map tensors to tensors, the $l^2$-norm is the most frequent choice. However, since FNOs map functions to functions, we can make use of the squared $L^2$-norm, which is also a much more appropriate choice when dealing with PDEs. It's worth noting the nuanced difference between the $L^2$-norm  and the $l^2$-norm, particularly when it comes to implementation. In both cases, you'll need to discretize the integral for computational purposes. However, the key distinction lies in the underlying space over which the norms are computed. When using the $L^2$-norm, one is essentially approximating the integral over a function space, aiming to capture the "true" behavior of the function. On the other hand, the $l^2$-norm is computed over an Euclidean space, essentially summing up the squared differences in a point-wise manner. Hence, the loss function for the training is:
\begin{align}
    L(\hat{G}, S_{\text{train}}) =  \frac{1}{N_s}\sum_{i=1}^{N_s} || \hat{G}(\boldsymbol{m}_i) - \boldsymbol{d}_i ||_{L^2}^2 + \lambda ||\boldsymbol{\theta}||_2^2, \quad (\boldsymbol{m}_i, \boldsymbol{d}_i) \in S_{\text{train}}.
\end{align}

 Both FNO and T-UNet show promise in approximating the high-fidelity forward model, ${G}$. T-UNet capitalizes on the strength of 2D convolutional layers and transformer architectures to efficiently incorporate temporal information and spatial parameters. Its design allows for an efficient encoding of multi-dimensional data, making it computationally more lightweight. On the other hand, FNOs offer resolution invariance and the capability to operate directly in function spaces, making them highly suitable for parametric PDE problems. However, one caveat with FNOs is their higher memory consumption. This is largely due to their need to include time as an additional channel and their use of 3D Fourier transforms and convolutions, which significantly increase the size of input tensors. 

The code for both the T-UNet and FNOs architectures is publicly accessible. The repository can be found at https://github.com/nmucke/subsurface-DA-with-generative-models.

\section{Data Assimilation }
\label{section:data_assimilation}
\subsection{Ensemble Smoother with Multiple Data Assimilation (ESMDA)}
\label{section:esmda}

Ensemble-based DA techniques are computationally efficient, as they lend themselves well to parallelization. Additionally, these methods offer a level of flexibility by requiring minimal alterations to the existing forward model code and avoiding the computation of adjoints gradients \cite{evensenDataAssimilationFundamentals2022}. Among various ensemble-based DA techniques, the Ensemble Smoother (ES) serves as an effective method but it has limited ability to provide adequate data matches in complex problems, such as reservoir simulations, due to the application of a single Gauss--Newton correction for conditioning the ensemble to all available data \cite{Emerick2013}. To address this, \citet{emerickEnsembleSmootherMultiple2013} introduced the ESMDA, an iterative version of ES, allowing for improved data matches by assimilating the same data multiple times with an inflated covariance matrix of measurement errors, which enables a more robust approach to updating the model. In spite of its inherent assumptions of Gaussianity, ESMDA is also applicable to weakly non-Gaussian problems. The method is easy to implement, leading to broad application in various scenarios.

Following the notation of \cite{Emerick2013}, we can describe the analysis for each ensemble member's model parameter \(\boldsymbol{m}_j^a\) as follows:
\begin{equation}
\boldsymbol{m}_j^a = \boldsymbol{m}_j^f + C^{f}_{MD} \left( C^{f}_{DD} + \alpha_i C_{D} \right)^{-1} \left( \boldsymbol{d}_{j} - G(\boldsymbol{m}_j)^f \right), \quad \text{for } j = 1, 2, \ldots, N_e.
\label{eq:esmda_update}
\end{equation}
Here \(N_e\) is the total number of ensemble members, \(\boldsymbol{m}_j^a\) and \(\boldsymbol{m}_j^f\) represent the analyzed and forecasted parameters of the \(j^{th}\) ensemble member, respectively, \(C^{f}_{MD}\) and \(C^{f}_{DD}\) are the cross- and auto-covariance matrices of the model parameters and data in the forecast step, \(\alpha\) is a scaling factor, and \(\boldsymbol{d}_{j}\) and \(G(\boldsymbol{m}_j)^f\) are the perturbed and forecasted observations for the \(j^{th}\) ensemble member. Besides, \(G()\) represents the forward model. For more details, see Equation \ref{eq:perturbed_obs}:
\begin{equation}
\boldsymbol{d}_{j} = \boldsymbol{d}_{\text{obs}} + \sqrt{\alpha} C^{1/2}_{D} z,
\label{eq:perturbed_obs}
\end{equation}
In this equation, \(\boldsymbol{d}_{\text{obs}}\) is the observed data, \(C_D\) is the measurement error covariance, and \(z\) is sampled from a standard normal distribution with zero mean and an identity matrix as the covariance. The factor \(\sqrt{\alpha}\) scales the perturbations.

The ESMDA algorithm, summarized below, iteratively employs these equations to update each ensemble member.
\begin{algorithm} \label{alg:online_stage}
\KwIn{Initial ensemble, observed data \(\boldsymbol{d}_{\text{obs}}\), and measurement error covariance \(C_D\)}
Determine \(N_a\) and \(\alpha\) for \(i = 1, \ldots, N_a\) \\
\For{$i = 1$ to $N_a$}{
    Compute predicted data \(G(\boldsymbol{m}_j)^f\) for each ensemble member \\
    Generate perturbed observations \(\boldsymbol{d}_{j}\) using Equation \ref{eq:perturbed_obs} \\
    Update ensemble members \(\boldsymbol{m}^a\) using Equation \ref{eq:esmda_update} \\    
}
\KwOut{Compute the final posterior results with DARTS with optimized parameters \(\boldsymbol{m}^j\)}.
\caption{ESMDA Algorithm}
\label{alg:esmda}
\end{algorithm}
In this study, the model parameter \(\boldsymbol{m}\) is defined as the permeability of the medium, and \(\boldsymbol{d}_{\text{obs}}\) represents pressure values obtained from specific monitoring points. 
The ESMDA methodology serves as the basis for a hybrid method that combines DA with surrogate modeling, with the objective of history matching in GCS studies.

\subsection{Randomized Maximum Likelihood (RML)}
\label{section:rml}

Randomized Maximum Likelihood is a variational DA method for approximating the posterior pdf with a method introduced in \citet{SOliver1996}. As a gradient-based method, it provides the advantage of better convergence and accuracy than an ensemble-based method, within a specified solution space. These benefits come at the cost of computational complexities. The need for adjoint models can be prohibitive and the computational efforts linked to the linearization of the model may introduce errors \cite{GarcaPintado2018}.

RML employs a set of cost functions, often denoted as \( J(\boldsymbol{m}_j) \). These cost functions aim to minimize the discrepancy between the ensemble's forecasted model states and the observed data, as well as the a priori model information. By optimizing these cost functions, RML generates multiple models that are consistent with the available observations, thus aiding in robust UQ.
The core idea of RML is to use a set of cost functions, \( J(\boldsymbol{m}_j) \), defined as:
\begin{equation}
J(\boldsymbol{m}_j) = ( \boldsymbol{m}_j - \boldsymbol{m}_j^{\text{prior}} )^T C_{MM}^{-1} ( \boldsymbol{m}_j - \boldsymbol{m}_j^{\text{prior}} ) +  ( G(\boldsymbol{m}_j) - \boldsymbol{d}_j )^T C_{DD}^{-1} ( G(\boldsymbol{m}_j) - \boldsymbol{d}_j ),
\label{eq:rml_objective_ensemble}
\end{equation}
where \( \boldsymbol{m}_j^{\text{prior}} \) and \( \boldsymbol{d}_j \) represent the prior model parameters and the perturbed observed data for the \( j^{th} \) ensemble member, respectively. These cost functions are designed to produce an ensemble of models that are coherent with the observed data, thereby assisting in robust DA.
For the minimization of the RML cost function \( J(\boldsymbol{m}_j) \), we employ the Adam optimizer due to its effectiveness and computational efficiency \cite{kingma2014adam}.

Applying RML in practice can be computationally intensive as each member of the ensemble necessitates a separate optimization process. Gradient-based optimization methods are commonly used for this purpose, and when adjoints are available for the forward model, they can be applied and increase efficiency.
Algorithm \ref{alg:rml} describes RML. It takes an initial set of model realizations and observations as inputs. It then computes the relevant covariance matrices and generates variations of the prior model and observed data for each ensemble member. Each ensemble member is then updated by optimizing its respective cost function.

\begin{algorithm} 
\label{alg:rml}
\KwIn{Initial model set \( \boldsymbol{m}_{\text{prior}} \) and perturbed observed data \( \boldsymbol{d}_j\)}
Compute covariance matrices \( C_{MM} \) and \( C_{DD} \) \\
Generate variations of the prior model \( \boldsymbol{m}_j^{\text{prior}} \) and perturbed observed data \( \boldsymbol{d}_j \) \\
\For{\( j = 1, \ldots, N_{\text{e}} \)}{
    Compute \( J(\boldsymbol{m}_j) \) using Equation \ref{eq:rml_objective_ensemble} \\
    Optimize \( J(\boldsymbol{m}_j) \) using gradient-based optimization (e.g. Adam)\\
    Compute the final posterior results with DARTS with optimized parameters \(\boldsymbol{m}^j\) \\    
}
\KwOut{Posterior history matched states for entire ensemble}
\caption{RML Algorithm}
\label{alg:rml}
\end{algorithm}

\section{Hybrid Data Assimilation}
\label{section:hybrib_DA}

\subsection{Surrogate-based hybrid ESMDA (SH-ESMDA)}
\label{section:hybrid_esmda}
ML surrogate models offer computational efficiency but may compromise robustness when used independently for DA. Conversely, ensemble-based DA methods like ESMDA are known for their accuracy but often come at a high computational cost. 
To preserve the trade-offs between efficiency and robustness, we introduce a surrogate-based hybrid approach called SH-ESMDA, which is within the ``Data Learning'' paradigm introduced by \citet{Buizza2022}. A key feature facilitating this integration is the use of the same parameters as input for both the ML algorithm and the reservoir simulator, allowing the surrogate model to serve as a direct substitute for the forward model in the intermediate steps of ESMDA. It is important to note that the surrogate model is only used in the intermediate steps of the ESMDA process. This ensures that the high-fidelity forward model is leveraged for both the initial and the computation of the posterior step, resulting in a more robust and accurate assimilation process. The surrogate model for SH-ESMDA requires training with only the number of forward simulations typically necessary for running the prior in standard approaches. This feature ensures no additional computational burden beyond conventional methods.

The development of SH-ESMDA has the primary objective of accelerating the ESMDA procedure, not achieving more accurate history-matching results than conventional methods. This is due to the fact that in this scheme, ESMDA is still the core DA method, so this hybrid approach also will keep its limitations in terms of DA. The main gain is the acceleration of the process, which potentially enables more iterations, which might otherwise be computationally prohibitive if relying solely on high-fidelity simulations. To achieve this acceleration, the following steps are proposed: 
\begin{enumerate}
  \item \textbf{Prior Dataset Generation}: Generate a prior dataset consisting of channelized permeability models with Alluvsim. Subsequently, perform CO\textsubscript{2} injection simulations on these models using the DARTS simulator.
  \item \textbf{Surrogate Model Training}: Train a surrogate model, such as a FNO or T-UNet, on the generated dataset. This training can be conducted as an offline stage, allowing the pretrained model to be reused in multiple subsequent Hybrid-ESMDA-Surrogate runs, thereby obviating the need for repetitive training and enhancing computational efficiency.
  \item \textbf{Initial Analysis Step}: Compute the first analysis step with ESMDA using prior forecasts.
  \item \textbf{Intermediate ESMDA Steps}: Employ the trained surrogate model as a substitute for DARTS in the intermediate steps of the ESMDA process. This offers a computationally efficient approximation to the solution.
  \item \textbf{Posterior}: Incorporate simulations from DARTS to compute the posterior to refine the solution and compute the final posterior states.
\end{enumerate}

The key innovation is leveraging the efficiency of the trained surrogate model to handle the computationally intensive ESMDA iterations. The final step with DARTS simulations acts as a physics-based regularizer to enhance robustness. Another significant advantage of a SH-ESMDA is the capability to pre-train the surrogate model in an offline stage. Once trained, this model can be reused across multiple SH-ESMDA runs without the need for retraining, thereby providing an additional layer of computational efficiency. This feature is particularly beneficial when dealing with a series of similar scenarios, as it eliminates the need to undergo the training process before each new ESMDA run. Consequently, this enables more frequent and rapid iterations, further enhancing the overall efficiency of the DA process.
This SH-ESMDA approach is summarized in the algorithm below:

\begin{algorithm}
\KwIn{Initial ensemble, observed data \(\boldsymbol{d}_{\text{obs}}\), measurement error covariance \(C_D\), and trained surrogate model}
Determine \(N_a\) and \(\alpha\) for \(i = 1, \ldots, N_a\) \\
\For{\(i = 1\) to \(N_{a}\)}{    
    \eIf{\(i = 1\)}{
        Compute the prior using DARTS \\
    }{
        Use surrogate model to compute predicted data \(\hat{G}(\boldsymbol{m}_j)^f\) for each ensemble member \\
    }
    Generate perturbed observations \(\boldsymbol{d}_{j}\) using Equation \ref{eq:perturbed_obs} \\
    Update ensemble members \(\boldsymbol{m}^a\) using Equation \ref{eq:esmda_update}
}
\KwOut{Compute the final posterior results with DARTS with optimized parameters \(\boldsymbol{m}^a\)}
\caption{Hybrid-ESMDA-Surrogate Algorithm}
\label{alg:hybrid_esmda}
\end{algorithm}

\subsection{Surrogate-based Hybrid RML (SH-RML)}
\label{section:hybrid_rml}

In a similar manner to SH-ESMDA, we introduce a surrogate-based hybrid (SH-RML), a method that integrates ML surrogates into the RML variational framework. One of the critical features that make this integration possible is the use of consistent permeability parameterization for both the surrogate and the high-fidelity DARTS model. This uniformity allows for seamless transitions between the surrogate and the physics-based models during the optimization process. SH-RML employs a streamlined approach similar to SH-ESMDA, where the surrogate model training occurs in the same fashion. It requires only as many forward simulations as are typically necessary in the prior phase of conventional methods. This ensures the method's computational efficiency by aligning with the standard simulation demands, and avoids the need for additional computational resources.

A particular feature of this method is its ability to allow variational DA, even in cases where simulators lack adjoint capabilities. This capability is achieved through the computation of gradients using automatic differentiation from a neural network. The primary objective of the SH-RML approach is to facilitate the optimization of the RML cost function \( J(\boldsymbol{m}_j) \). The algorithmic flow of the SH-RML is as follows:

\begin{enumerate}
  \item \textbf{Prior Dataset Generation}: Generate a prior dataset consisting of channelized permeability models with Alluvsim. Subsequently, perform CO\textsubscript{2} injection simulations on these models using the DARTS simulator.
  \item \textbf{Surrogate Model Training}: Train a surrogate model, such as a FNO or T-UNet, on this dataset. This training can be conducted as an offline stage, allowing the pre-trained model to be reused in multiple subsequent SH-RML runs.
  \item \textbf{Parameter Initialization}: Initialize the permeability parameters \( \boldsymbol{m}_j \) for each ensemble member within the RML framework.
  \item \textbf{Initial Optimization}: Use gradients derived from the surrogate model to perform initial optimization of the cost function \( J(\boldsymbol{m}_j) \).
  \item \textbf{Posterior Parameter Computation}: Compute the posterior parameters after the surrogate-based optimization using RML.
  \item \textbf{High-Fidelity Refinement}: Apply these posterior parameters to the DARTS model, running high-fidelity simulations to refine the solution and compute the final posterior states.
\end{enumerate}

The initial steps of the RML optimization process are accelerated by leveraging the surrogate model, which offers both efficiency and automatic differentiation capabilities. The final steps employ the DARTS model to ensure high-fidelity, physics-based solutions as summarized in the algorithm below:

\begin{algorithm} 
\KwIn{Initial model set \( \boldsymbol{m}_{\text{prior}} \), perturbed observed data \( \boldsymbol{d}_j \), and trained surrogate model}
Compute covariance matrices \( C_{MM} \) and \( C_{DD} \) \\
Generate variations of the prior model \( \boldsymbol{m}_j^{\text{prior}} \) and perturbed observed data \( \boldsymbol{d}_j \) \\
\For{\( j = 1, \ldots, N_{\text{e}} \)}{
    Use surrogate to compute initial \( J(\boldsymbol{m}_j) \) and gradients \\
    Perform optimization of \( J(\boldsymbol{m}_j) \) using gradient-based methods \\    
    }
\KwOut{Compute the final posterior results with DARTS with optimized parameters \( \boldsymbol{m}_j \) \\}
\caption{Hybrid-RML-Surrogate Algorithm}
\label{alg:hybrid_rml}
\end{algorithm}

One of the key innovations in SH-RML is the utilization of automatic differentiation capabilities provided by the surrogate model. This eliminates the need for manually deriving computationally expensive adjoint models, which are required in traditional variational DA methods. As a result, the SH-RML offers an efficient approach to DA in complex, nonlinear systems. However, it is important to acknowledge the inherent limitations due to the intrinsically ill-posed nature of the problem, affecting the overall results.

In summary, the proposed SH-ESMDA and SH-RML frameworks offer solutions for enhancing DA techniques in applications such as GCS. By adding the computational advantages of ML surrogates with the reliability of physics-based models, these hybrid methods pave the way for efficient and accurate DA and, ultimately, a better understanding and quantification of the model and data uncertainties.

\section{Results}
\label{section:results}
In this section, we present a comprehensive evaluation of surrogate models and history-matching methods for GCS applications. We begin with the training and evaluation of FNO and T-UNet as surrogate models. This is followed by an assessment of the ESMDA method for history matching. Subsequently, we evaluate the SH-ESMDA. Finally, we discuss the SH-RML results. 

\subsection{Training and evaluation of the FNO and T-UNet}
\label{sec: NN_traintest}
In this Section, the focus is on training the T-UNet and the FNO to serve as surrogate models that approximate the high-fidelity forward model $G$. As described in Section \ref{section:neural_networks}, these models are trained on a dataset derived from high-fidelity reservoir simulations generated with the DARTS simulator. This dataset includes critical unknowns such as permeability, porosity, and gas injection rates, along with corresponding state variables, pressure and CO\textsubscript{2} molar fraction.  Given the developing state of understanding of optimal FNO configurations, we particularly investigated the impact of varying Fourier modes, considering both modes 18 and 6, and hidden channel widths of 128 and 64. This nuanced examination aims to contribute to the open question of how to best configure FNOs for subsurface modeling tasks.

\begin{figure}[h!]
    \centering
    \begin{subfigure}{0.8\textwidth}
        \includegraphics[width=\textwidth]{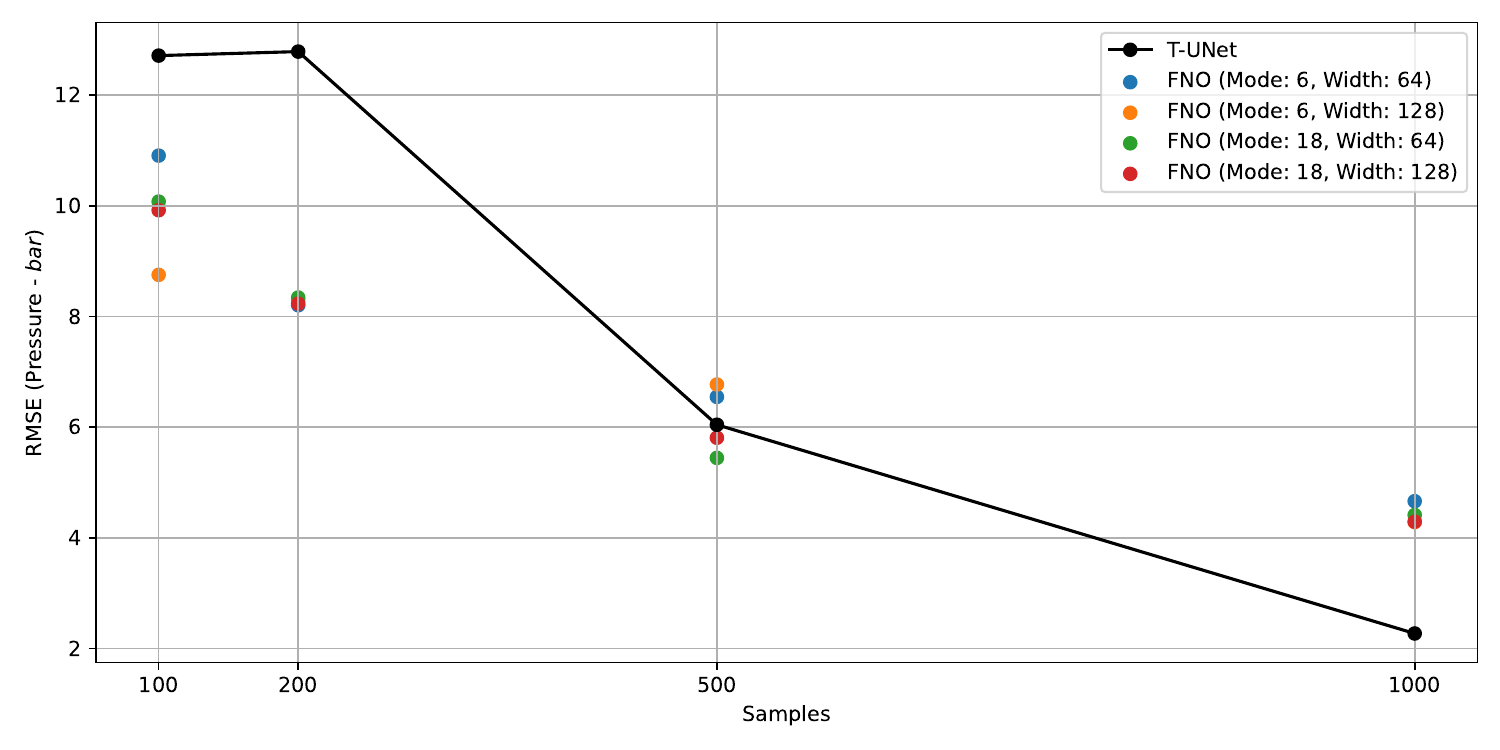}
        \caption{Test RMSE Metrics for Pressure }
        \label{fig:pressure}
    \end{subfigure}
    \hfill
    \begin{subfigure}{0.8\textwidth}
        \includegraphics[width=\textwidth]{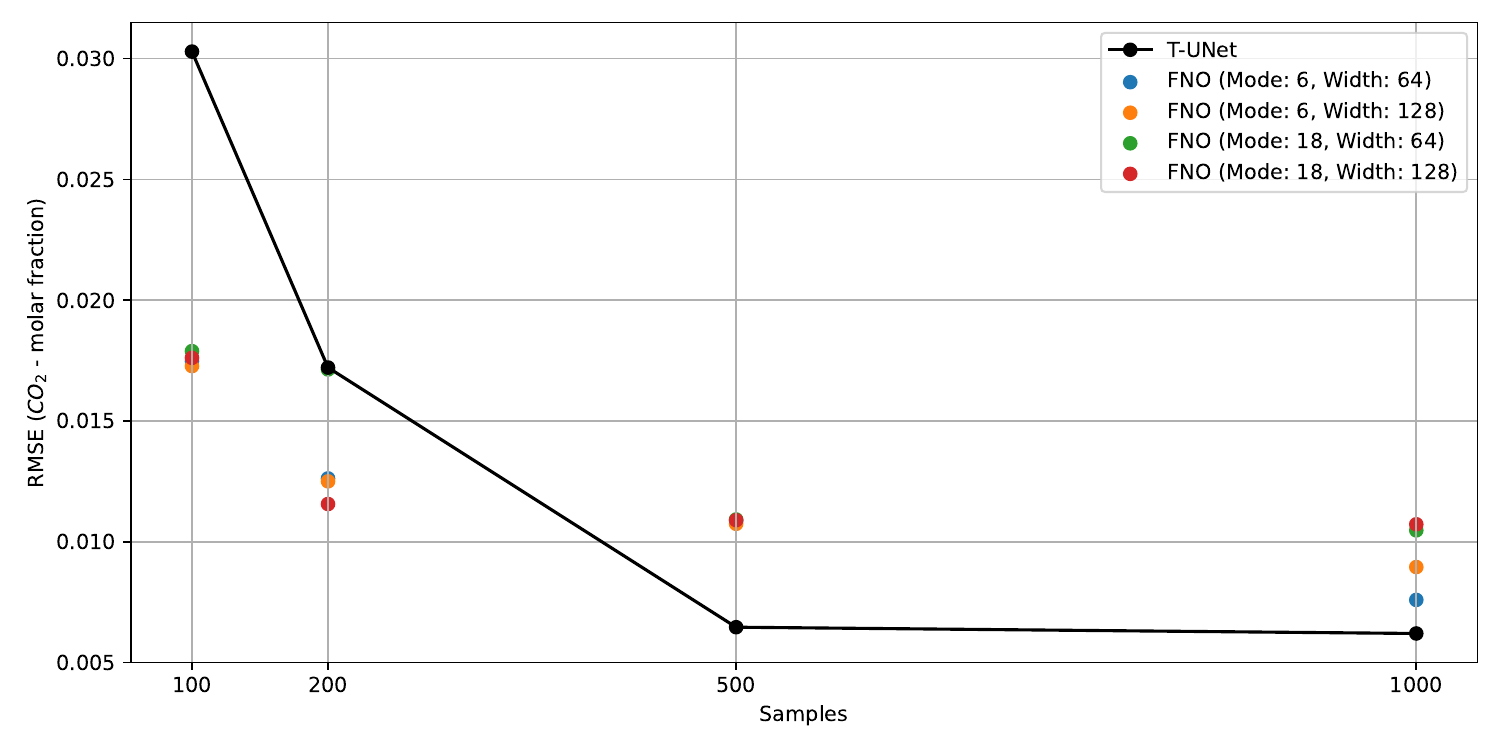}
        \caption{Test RMSE Metrics for \( \text{CO}_2 \) molar fraction}
        \label{fig:CO2}
    \end{subfigure}
    \caption{Test RMSE Metrics for FNO and T-UNet}
    \label{fig:rmse_metrics}
\end{figure}

To quantify the performance of the T-UNet and FNO surrogate models, we RMSE as a metric, focusing on both pressure and CO\textsubscript{2} molar fraction. Specifically, we consider training plus test sample sizes of 100, 200, 500, and 1000 to understand how the size of the training set affects the model's performance. Training and test data are split into 80\% and 20\%, respectively. The RMSE values presented as a function of the number of training samples in Figure~\ref{fig:rmse_metrics}, and detailed in the \ref{sec: appendix}, indicate that both neural network architectures yield accurate approximations of the high-fidelity forward model $G$. However, the FNO shows a slight advantage when the number of training samples is limited, particularly at the 100-sample size. This is a significant observation for subsequent data assimilation studies, as it suggests that FNO may require fewer training samples than the T-UNet, thus alleviating the need for additional high-fidelity simulations for neural network training. For context, it's important to note that the pressure range in the reservoir simulations is between 200 and 320 bars, and the CO\textsubscript{2} molar fraction varies from 0 to 1. In this range, the RMSE values indicate that the approximation errors are significantly small. However, these are still approximations and, although they are highly accurate, they can’t entirely replicate the high-quality DARTS simulations.

\begin{figure}[h!]
    \centering
    \begin{subfigure}{0.32\textwidth}
        \includegraphics[width=\textwidth]{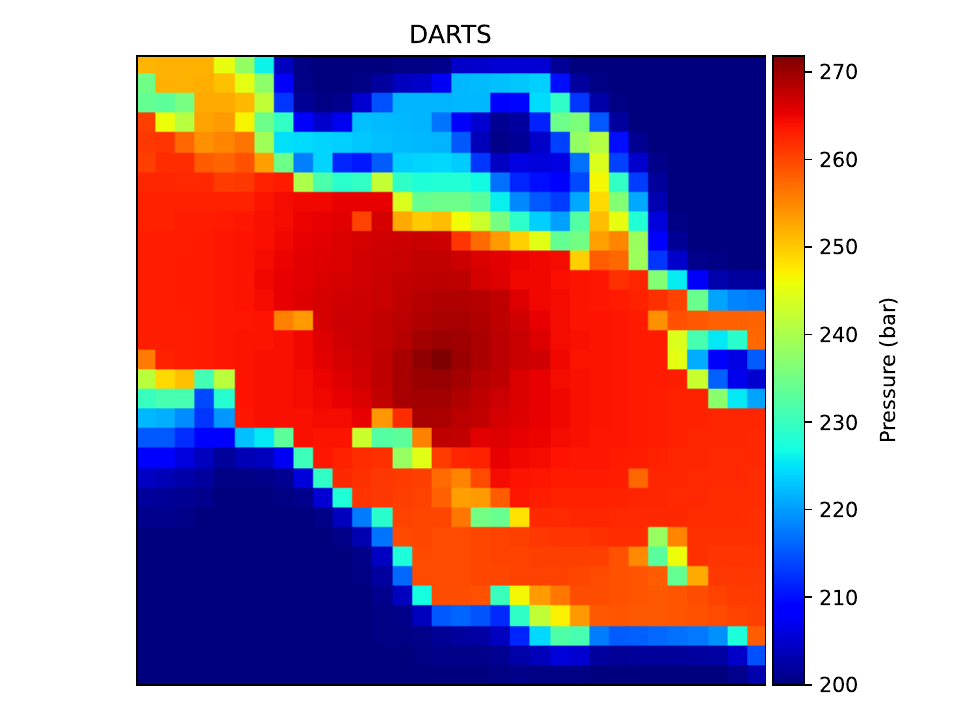}
        \caption{DARTS Pressure}
        \label{fig:true_pressure}
    \end{subfigure}
    \hfill
    \begin{subfigure}{0.32\textwidth}
        \includegraphics[width=\textwidth]{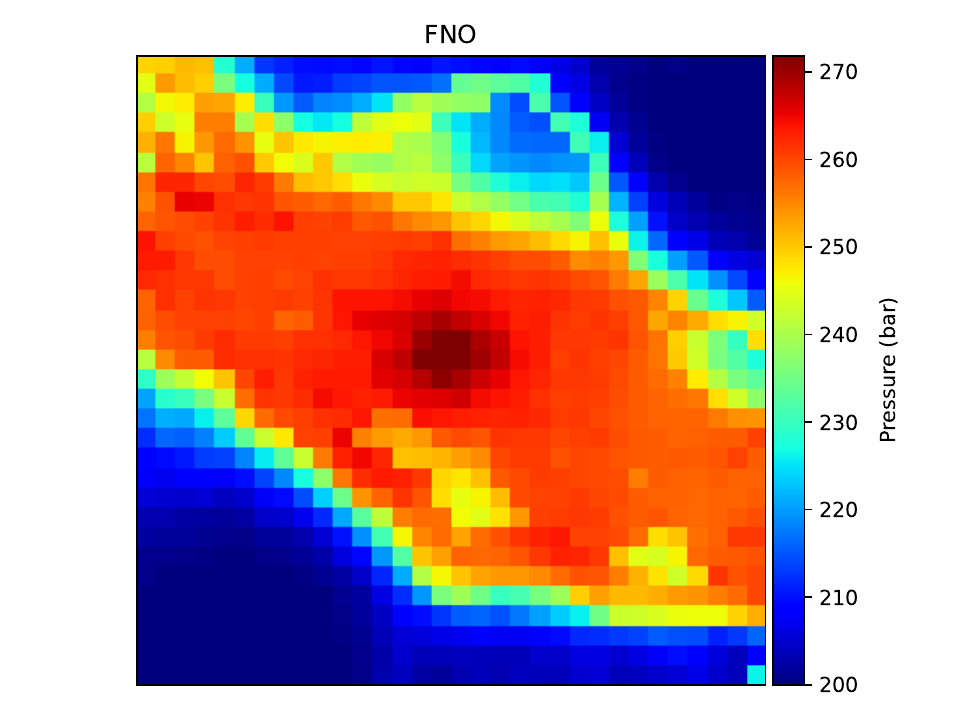}
        \caption{FNO Predicted Pressure}
        \label{fig:fno_pressure}
    \end{subfigure}
    \hfill
    \begin{subfigure}{0.32\textwidth}
        \includegraphics[width=\textwidth]{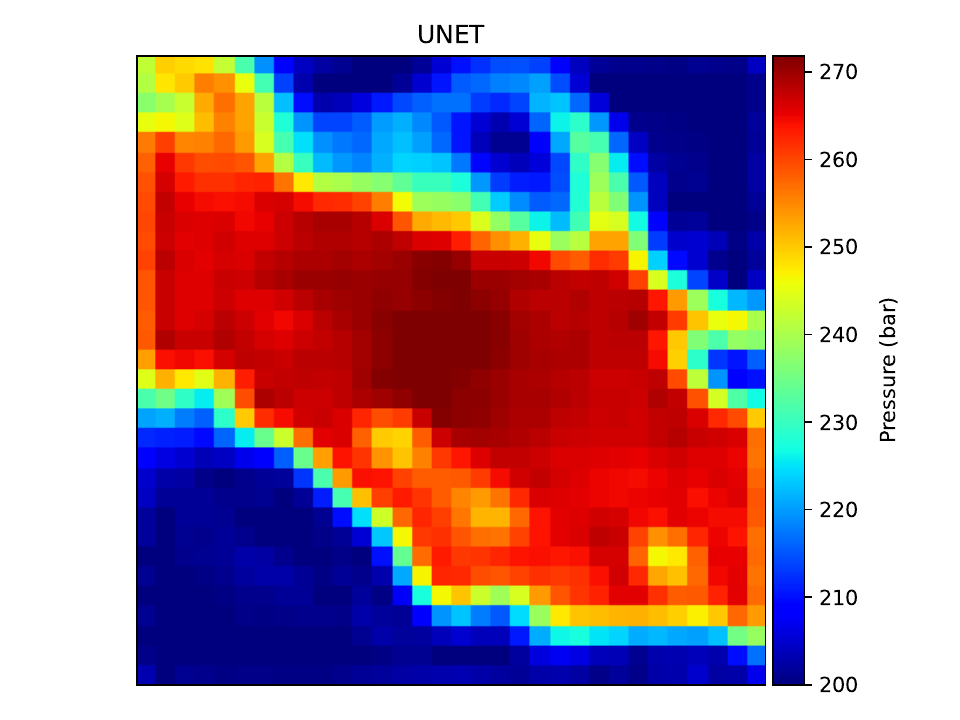}
        \caption{T-UNet Predicted Pressure}
        \label{fig:tunet_pressure}
    \end{subfigure}
    \caption{Pressure Distributions for a test case}
    \label{fig:pressure_distributions}
\end{figure}

\begin{figure}[h!]
    \centering
    \begin{subfigure}{0.32\textwidth}
        \includegraphics[width=\textwidth]{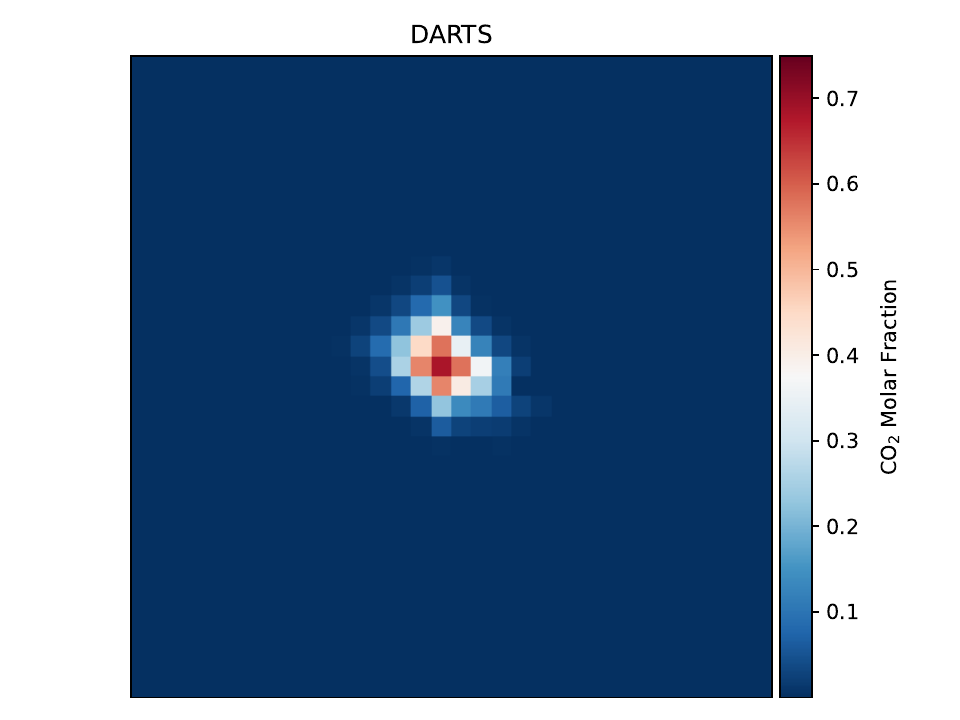}
        \caption{DARTS \( \text{CO}_2 \) molar fraction}
        \label{fig:true_CO2}
    \end{subfigure}
    \hfill
    \begin{subfigure}{0.32\textwidth}
        \includegraphics[width=\textwidth]{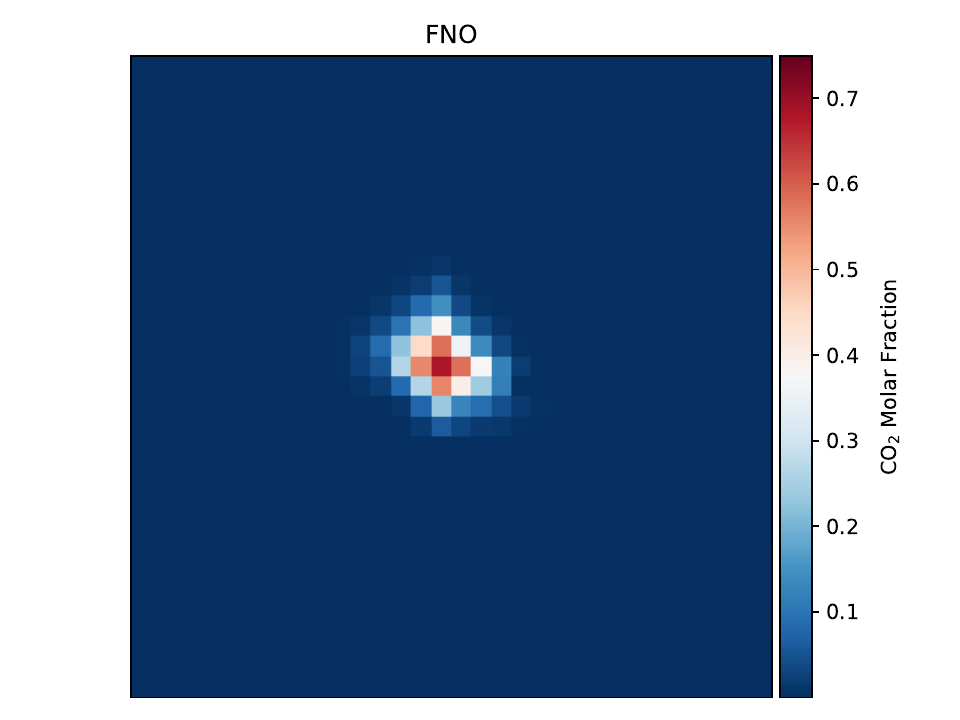}
        \caption{FNO Predicted \( \text{CO}_2 \) molar fraction}
        \label{fig:fno_CO2}
    \end{subfigure}
    \hfill
    \begin{subfigure}{0.32\textwidth}
        \includegraphics[width=\textwidth]{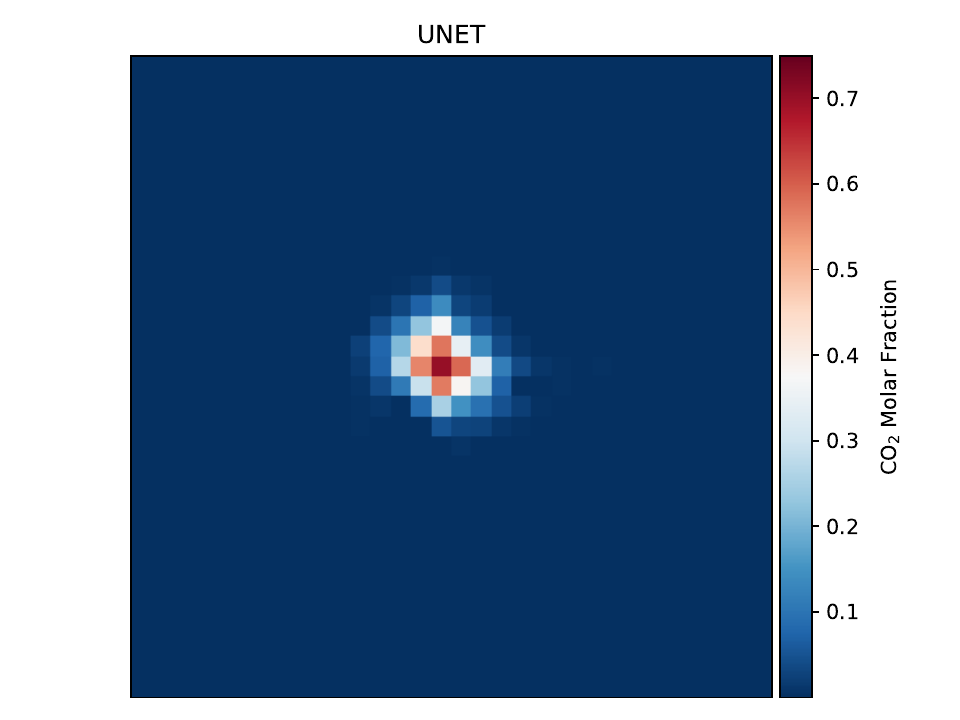}
        \caption{T-UNet Predicted \( \text{CO}_2 \) molar fraction}
        \label{fig:tunet_CO2}
    \end{subfigure}
    \caption{\( \text{CO}_2 \) distributions for a test case}
    \label{fig:CO2_distributions}
\end{figure}

\begin{figure}[h!]
    \centering
    \begin{subfigure}{0.45\textwidth}
        \includegraphics[width=\textwidth]{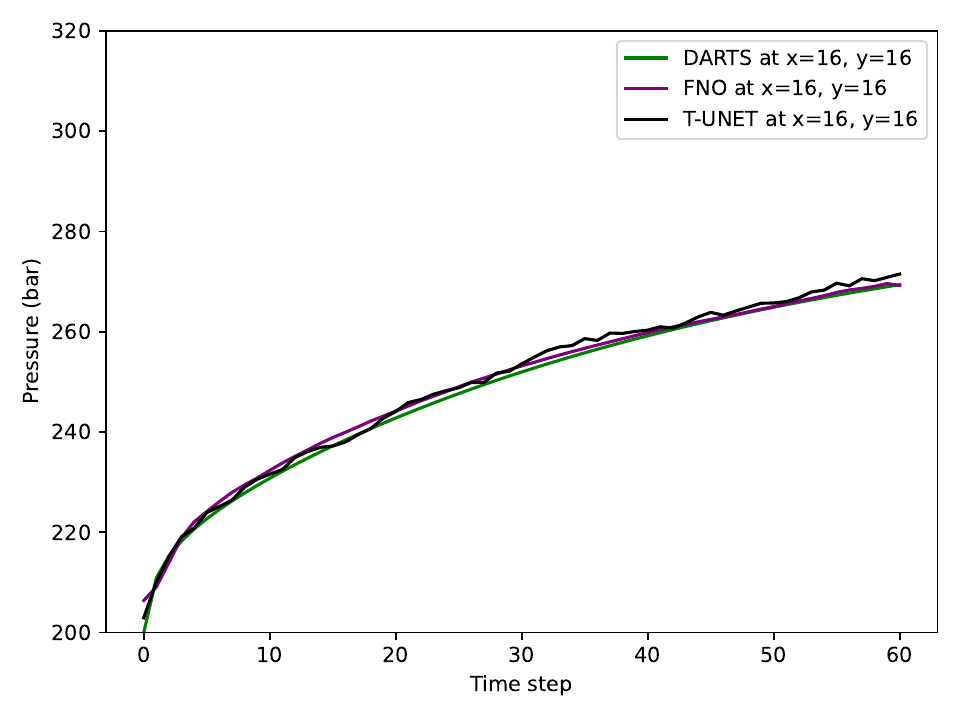}
        \caption{Time evolution of Pressure at grid (16,16)}
        \label{fig:pressure_time_evolution}
    \end{subfigure}
    \hfill
    \begin{subfigure}{0.45\textwidth}
        \includegraphics[width=\textwidth]{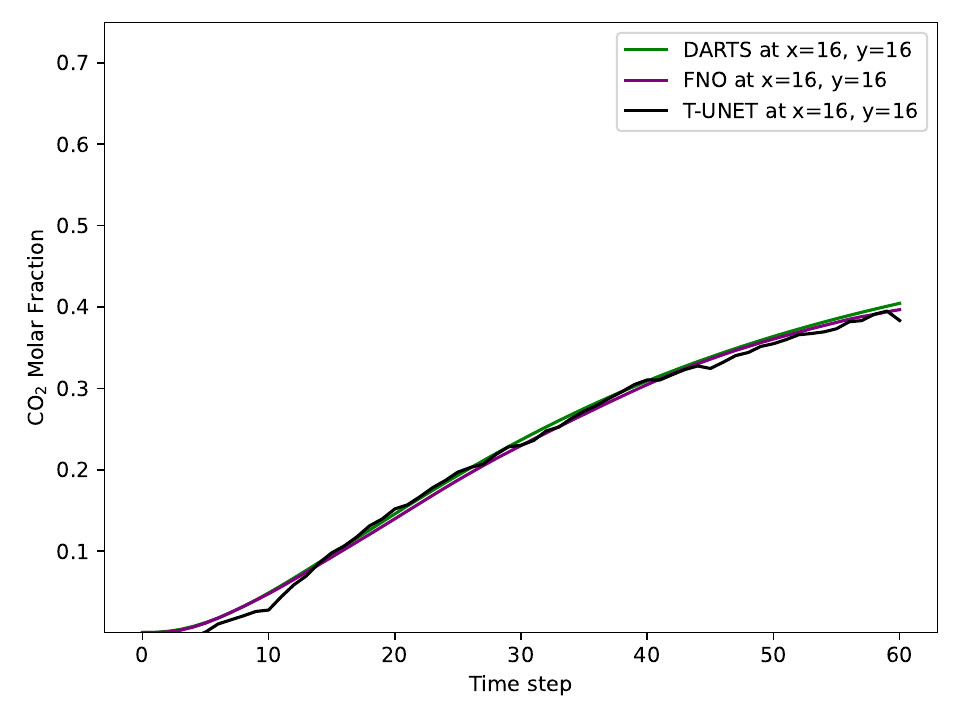}
        \caption{Time evolution of CO\textsubscript{2} molar fraction at grid (16,16)}
        \label{fig:CO2_time_evolution}
    \end{subfigure}
    \caption{Comparison of time evolution of states at grid (16,16) for DARTS, FNO and the T-UNet}
    \label{fig:NN_time_evolution}
\end{figure}

We also examine the capability of FNO and T-UNet to represent the spatial distributions of pressure and CO\textsubscript{2} molar fraction. Figures \ref{fig:pressure_distributions} and \ref{fig:CO2_distributions} present the contour maps of CO\textsubscript{2} distribution for a specific test case (not part of the training dataset). At the final time step, we compare the true state variables against those predicted by FNO and T-UNet. Those are responses for the models with a sample size of 1000. Both models capture the overall behavior of the reservoir and what's particularly noteworthy is the ability of both models to approximate the shape of the  CO\textsubscript{2} plume. We analyze the models' time evolution in addition to comparing their spatial distributions of pressure and CO\textsubscript{2} molar fraction at a specific point in the grid. Figure \ref{fig:NN_time_evolution} illustrates the temporal variations at the injection point located at the grid position (16,16). These figures show the predictive capabilities of both FNO and T-UNet in capturing dynamic behavior, as compared to the high-fidelity DARTS simulations. One observation that stands out is the relative smoothness in the time evolution generated by the FNO model, especially when compared to the more fluctuating curves from the T-UNet model. This difference is intriguing, and it prompts further discussion.

The FNO's less noisy, and thus more physically plausible, representation of the temporal evolution at the injection point may be attributed to the way it handles time steps as additional channels of data. This treatment allows the FNO to account for interdependencies across time, thus yielding a smoother, more coherent dynamic response. On the other hand, the T-UNet model generates time evolution independently from each other through integration with the transformer, as discussed in Section \ref{section:neural_networks}. This approach seems to result in a more noisy representation of the pressure, possibly because the temporal dependencies are not as explicitly captured as in the FNO model.

The difference in noise levels between the FNO and T-UNet models might not just be a matter of the numerical accuracy of the methods, but could also have implications for their respective usefulness in subsurface modeling and DA studies. The smoother temporal response of the FNO model may make it more suitable for cases that require a higher accuracy for the physical representation of the states. In summary, the ability of these models to adequately represent both the spatial and temporal complexities of the reservoir suggests their robustness and reliability for further analysis and DA studies. However, the observed differences in dynamic behavior between the FNO and T-UNet models could be a crucial factor in deciding which model to employ for specific applications.

\subsection{ESMDA History Matching}

\begin{figure}[ht!]
\centering
\includegraphics[width=0.5\linewidth]{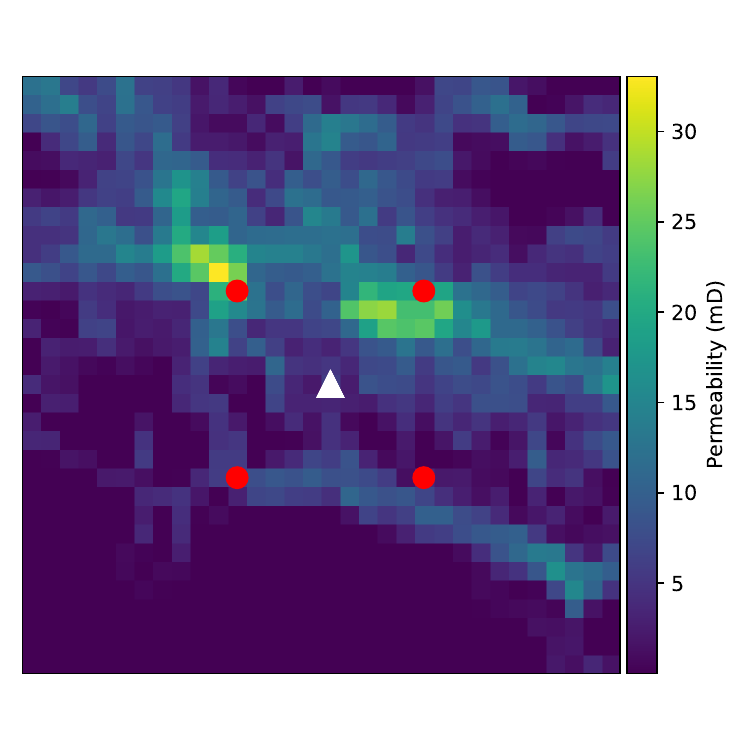}
\caption{Reference permeability model showcasing the central injection well (triangle) and peripheral monitoring points (circles). Monitoring points are numbered from 1 to 4, starting from the top-left corner and proceeding clockwise.}
\label{fig:ref_model}
\end{figure}

ESMDA is employed to conduct history matching on our reservoir model described in section \ref{section:reservoir_modeling}. The primary objective is to evaluate ESMDA's efficiency and accuracy in modeling complex reservoir systems for this GCS application. The task at hand poses challenges due to the nonlinearity of the problem and the non-Gaussian nature of the reservoir's permeability field. To start our assessment, we utilized a distinct reference permeability model, generated outside our prior distribution, to produce synthetic observed data. This model, displayed in Figure~\ref{fig:ref_model}, incorporated a central injection well, surrounded by four pressure monitoring points. 

\begin{figure}[ht!]
\centering
\includegraphics[width=0.85\linewidth]{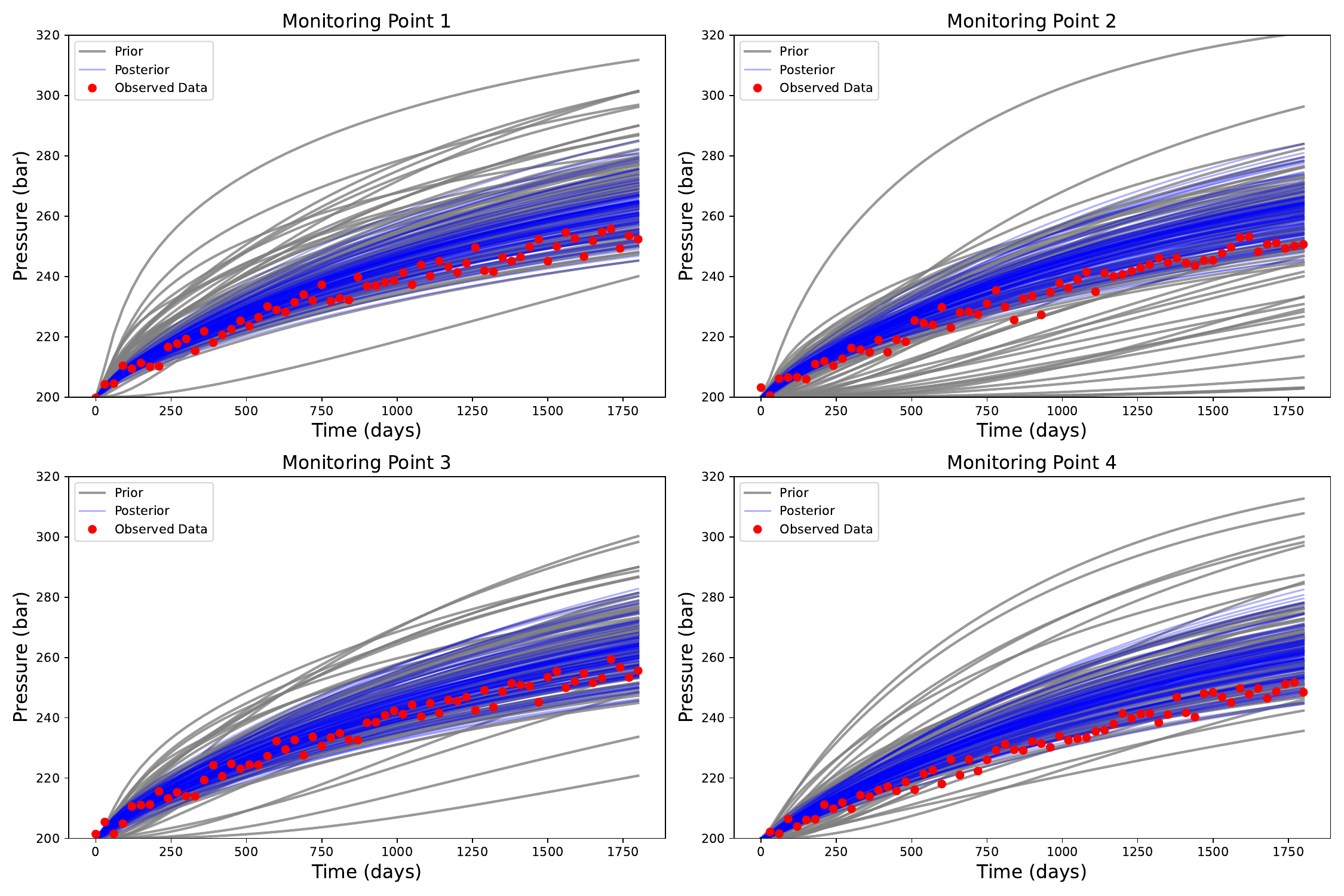}
\caption{Comparison between the prior, reference model, and posterior pressures at each monitoring point for the ESMDA history matching. Red dots represent a realization of perturbed observed data.}
\label{fig:esmda_pmatch}
\end{figure}

\begin{figure}[ht!]
\centering
\includegraphics[width=0.5\textwidth]{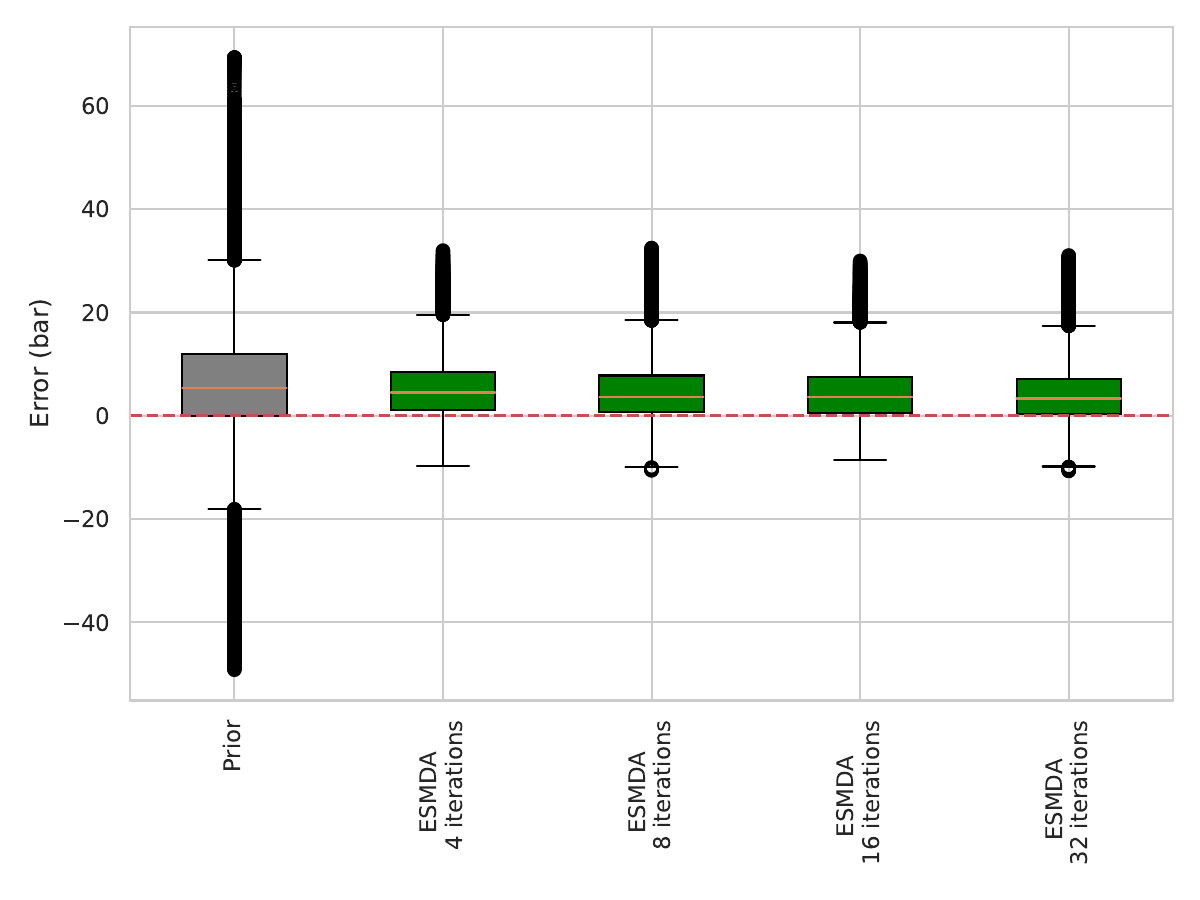}
\caption{Comparison of monitoring pressure absolute error across different ESMDA iterations and the prior. Red dots represent a realization of perturbed observed data.}
\label{fig:esmda_error}
\end{figure}

The ESMDA approach utilizes an ensemble of 100 prior permeability maps to represent the uncertainty before DA. These maps are drawn from geological realizations using the Alluivsim algorithm. Over 4 ESMDA steps, the results indicate that the posterior pressure distribution at each of the four monitoring points is closer to the observed data compared to the prior. Figure~\ref{fig:esmda_pmatch} displays this improvement, comparing pressures from the prior, the reference model, and the posterior. We further analyze the sensitivity of the method to the number of steps in ESMDA at the monitoring pressure points for both the prior and posterior models by varying the number of ESMDA steps: 4, 8, 16, and 32. Figure \ref{fig:esmda_error} reveals that increasing the number of iterations does not substantially improve the quality of history matching.

\begin{figure}[ht!]
\centering
\includegraphics[width=0.85\linewidth]{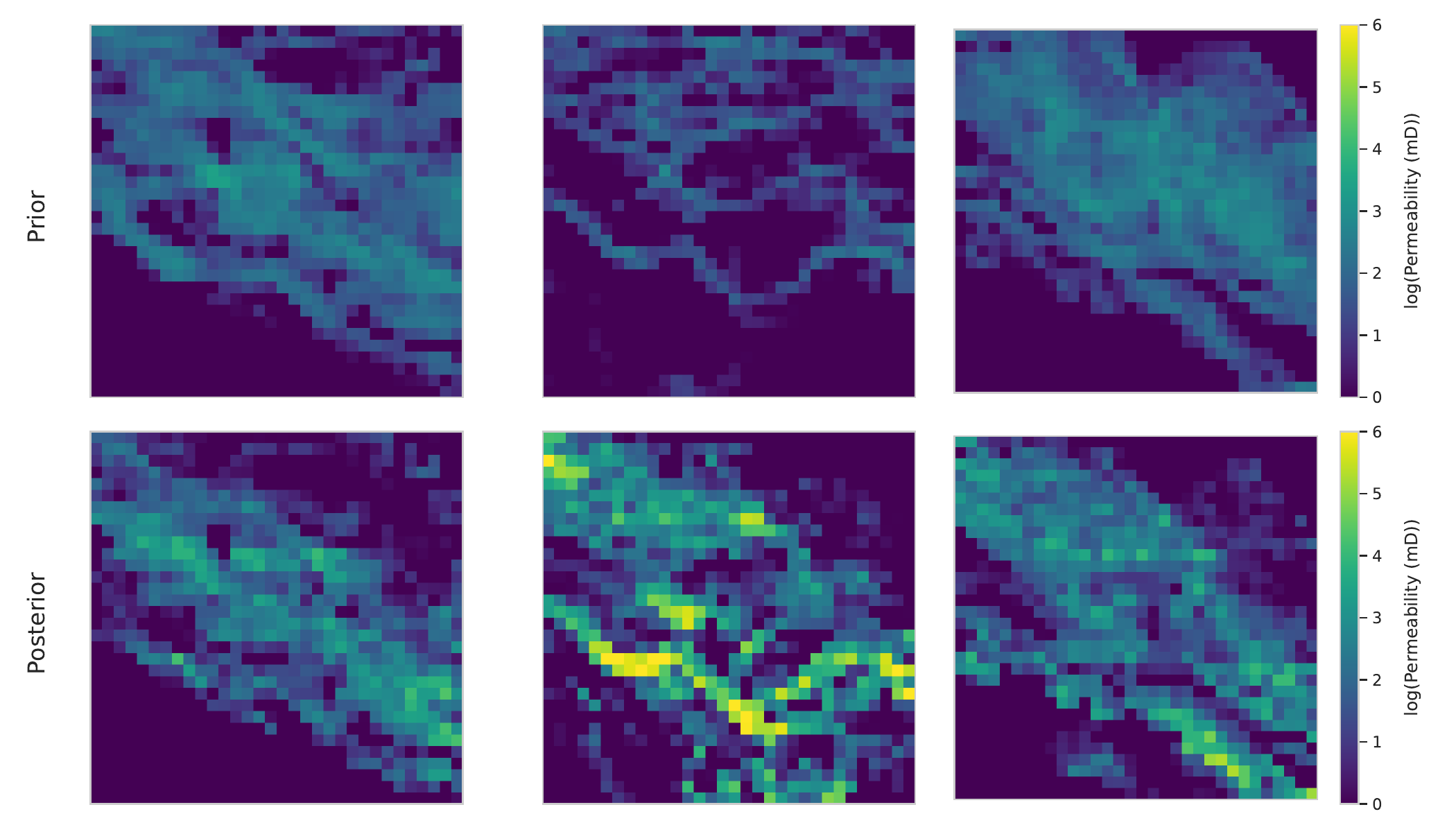}
\caption{Comparison between prior and posterior permeabilities across three different samples for the ESMDA.}
\label{fig:esmda_perm}
\end{figure}

Although ESMDA reduces errors related to measured pressure in comparison to the prior, it significantly overestimates reservoir permeability in comparison to prior permeability distributions, as illustrated in Figure \ref{fig:esmda_perm}. While undesirable, the discrepancy can be explained by the fact that history matching is an ill-posed problem, allowing for multiple solutions that can satisfactorily fit the data.

Given the inherent assumptions of ESMDA, namely, its reliance on Gaussian distributions and its better suitability for linear problems—none of which are present in our case—it is crucial to recognize its limitations in addressing the ill-posed problems we encounter. Our subsequent results delve into hybrid methods, as detailed in Section~\ref{section:hybrib_DA}. These techniques serve dual purposes: one aims to accelerate the computational process, while the other focuses on enhancing the accuracy of history matching.

\subsection{Results for SH-ESMDA}

\begin{figure}[ht!]
\centering
\includegraphics[width=1\linewidth]{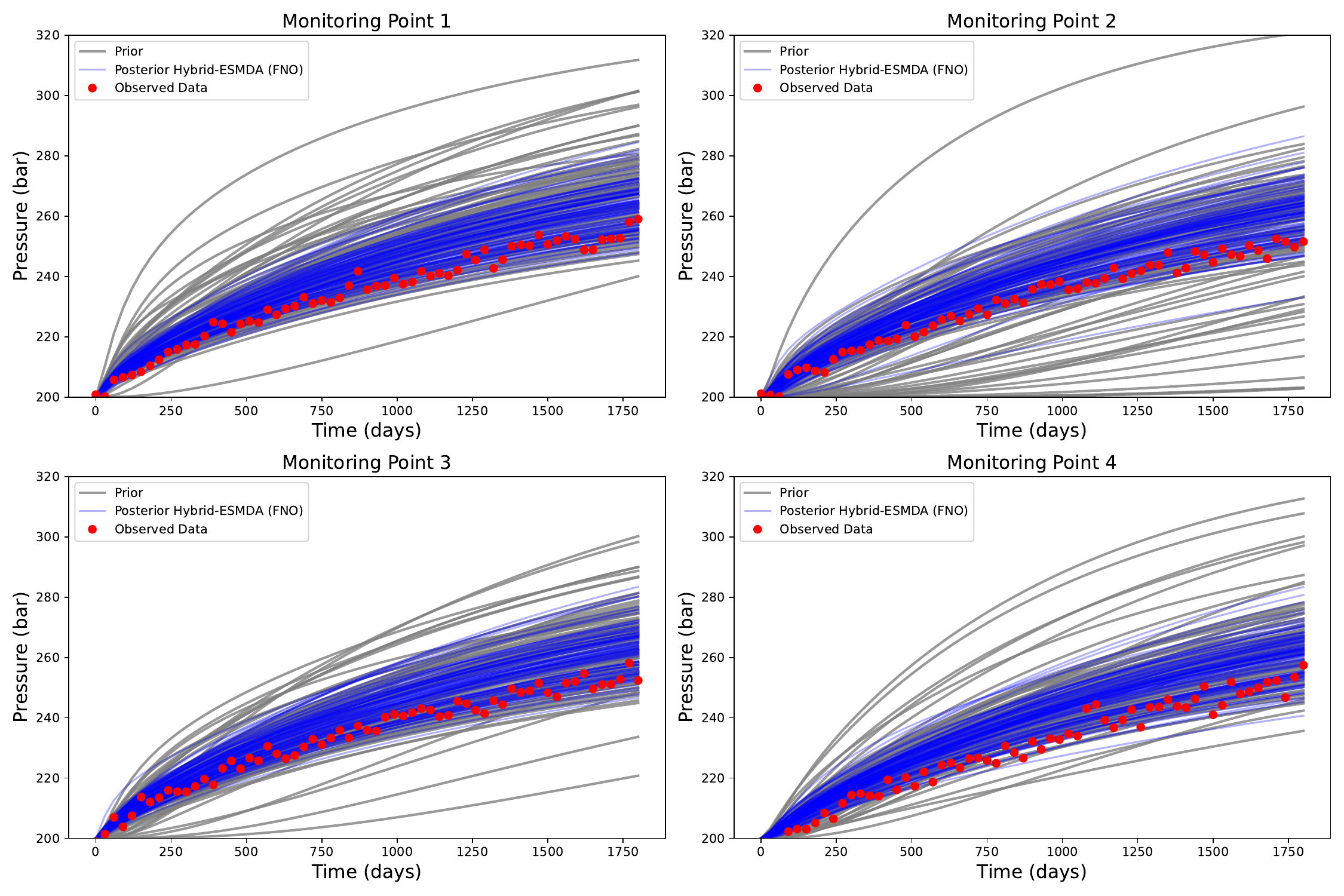}
\caption{Comparison between the prior, reference model, and posterior pressures at each monitoring point for the Hybrid-ESMDA-Surrogate history matching using FNO. Red dots represent a realization of perturbed observed data.}
\label{fig:fno_esmda_pmatch}
\end{figure}

\begin{figure}[h!]
\centering
\includegraphics[width=1\linewidth]{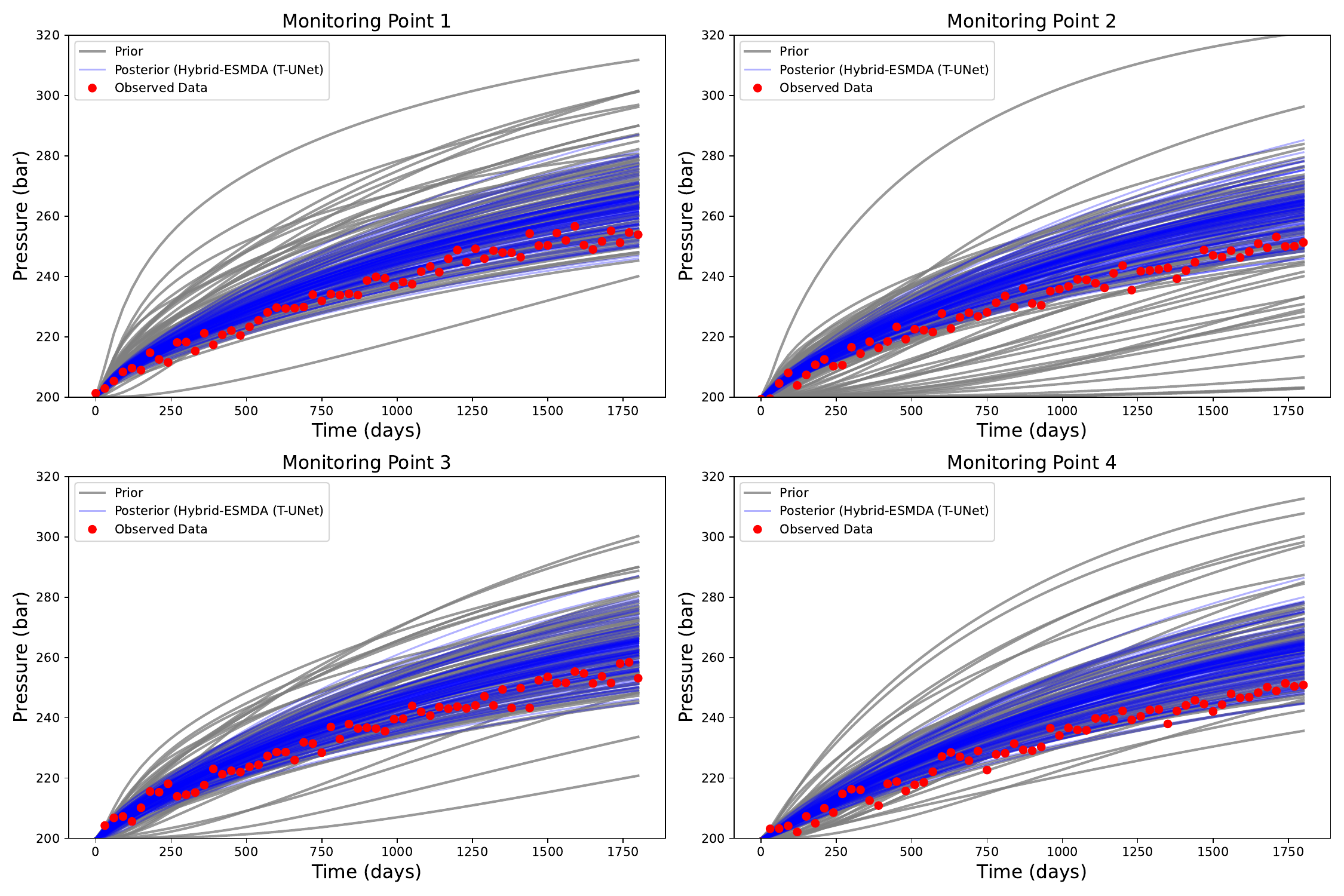}
\caption{Comparison between the prior, reference model, and posterior pressures at each monitoring point for the SH-ESMDA history matching using T-Unet. Red dots represent a realization of perturbed observed data.}
\label{fig:unet_esmda_pmatch}
\end{figure}

In order to accelerate the computational efficiency of the standard ESMDA, we employ ML surrogates in the form of FNO and T-UNet following the algorithm outlined in Section \ref{section:hybrid_esmda}. For this study, we employ 100 samples and conduct history matching for pressure at four monitoring points over four iterations. Figures~\ref{fig:fno_esmda_pmatch} and \ref{fig:unet_esmda_pmatch} display the results of these history-matching exercises. In this context, it should be highlighted that the larger-scale training experiments involving additional models were designed for performance benchmarking of SH-ESMDA method. These figures reveal the performance when FNO and T-UNet are used as surrogate models for the intermediate steps in the ESMDA algorithm, respectively. As can be observed, the outcomes produced by both FNO and T-UNet are remarkably similar to each other and closely align with those obtained using the standard ESMDA methodology with 4 steps.

\begin{figure}[ht!]
\centering
\includegraphics[width=0.5\textwidth]{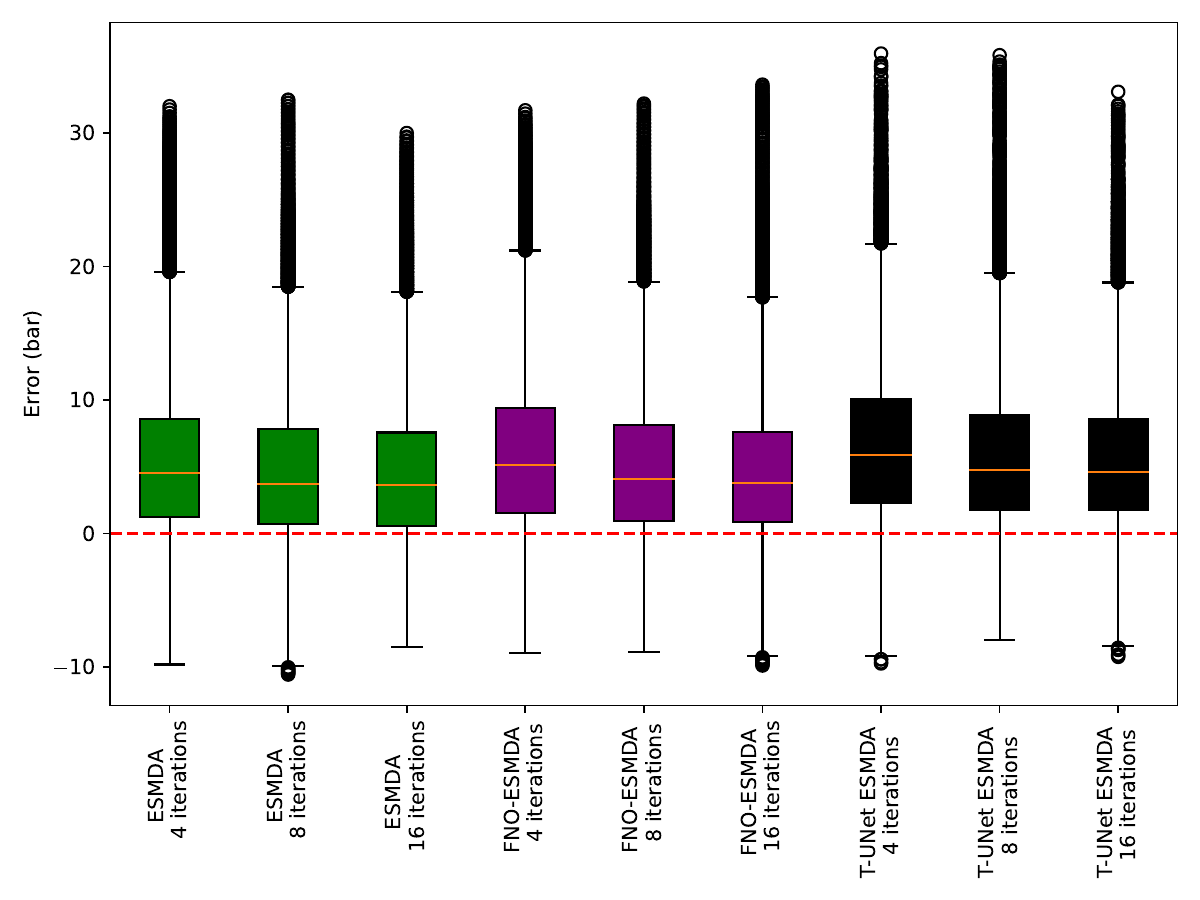}
\caption{Box plots comparing monitoring pressure errors between standard ESMDA (green) and SH-ESMDA using FNO (purple) and T-UNet (black) surrogates for 4, 8 and 16 iterations.}
\label{fig:hesmda_error}
\end{figure}

Figure~\ref{fig:hesmda_error} presents box plots that offer a comparison of the monitoring pressure errors for both the standard ESMDA and the enhanced SH-ESMDA methodology, employing FNO and T-UNet as surrogates. The SH-ESMDA methods maintain a level of accuracy that is comparable to that of the standard ESMDA, while significantly reducing the computational time required. When the T-UNet model is employed as the surrogate, the errors observed are marginally higher compared to when the FNO is used. It is also crucial to highlight that, consistent with the observations made for the standard ESMDA, increasing the number of iterations does not result in a substantial reduction in errors. 

The gains in computational efficiency are enumerated in Table \ref{tab:hesmda_time}. As the table illustrates, both versions of the SH-ESMDA method—utilizing either FNO or T-UNet as surrogates significantly reduce computation time. Specifically, there is a minimum speedup of around 50\%. In all experiments, the ESMDA involved  4 steps. The computational times reported in Table~\ref{tab:hesmda_time} do not include the time required to train the FNO and T-UNet models. The exclusion is justified for two main reasons. First, the training of these neural network models is generally considered a one-time computational expense. For the configuration used in our hybrid methods—comprising 100 samples for both FNO and T-UNet, along with 6 models and a width of 64—both models can be trained in less than 30 minutes. Once trained, these models can be reused for multiple iterations or different scenarios without the need for retraining. This is particularly beneficial in applications where the same or similar systems are studied multiple times. Second, the training can be performed offline and in parallel, taking advantage of high-performance computing resources. This minimizes its impact on the overall computational efficiency when amortized over multiple applications.

\begin{table}[h!]
  \centering
  \caption{Comparison of computational times for different methods, considering 100 ensemble members.}
  \label{tab:hesmda_time}
  \begin{tabular}{cccc}
    \toprule
    \multirow{2}{*}{Total number of steps} & \multicolumn{3}{c}{Methods} \\
    \cmidrule(lr){2-4}
    & ESMDA & Hybrid-ESMDA-Surrogate (FNO) & Hybrid-ESMDA-Surrogate (T-UNet) \\
    & (min) & (min) & (min) \\
    \midrule
    4  & 73  & 39 & 38 \\
    8  & 149 & 46 & 44 \\
    16 & 303 & 58 & 55 \\
     \bottomrule
  \end{tabular}
\end{table} 

The posterior permeability distributions obtained through both the SH-ESMDA schemes employing FNO and T-UNet closely align with those achieved using the standard ESMDA approach. Given that ESMDA serves as the core method for DA in these hybrid algorithms, the permeability distributions inherently reflect the strengths and limitations of ESMDA itself. Therefore, the hybrid schemes do not necessarily bring about a qualitative shift in the outcome; rather, their primary advantage lies in computational speedup. We acknowledge that the hybrid methods will inherit the constraints of ESMDA, and its inadequacies in encapsulating the uncertainties inherent in the system. For brevity and to avoid redundancy, we refrain from presenting additional figures showcasing the similarity in the posterior permeability distributions across the different methods, as they would closely mirror the results already discussed for ESMDA.

In summary, SH-ESMDA successfully combines the computational efficiency of ML with the reliability of DARTS like ESMDA. The approach reduces the time needed for each computational step, allowing for less time for the same amount of iterations. However, the SH-ESMDA inherits the fundamental weaknesses of the core ESMDA algorithm. Therefore, although the method provides substantial acceleration, the accuracy of the history matching is still constrained by the inherent limitations of ESMDA. This makes this SH-ESMDA scheme an important tool for accelerating DA studies, especially for complex reservoir models. 

\subsection{Results for SH-RML}

As elaborated in Section \ref{section:hybrid_rml}, SH-RML offers a combination of factors that make it an effective choice for accurate DA for complex problems. Here, we explore further its comparative results over other methods like ESMDA. Our findings show that the SH-RML improves history matching and provides more accurate posterior permeabilities estimates. For a comprehensive perspective on the computational requirements of SH-RML, it's pertinent to discuss the computational time in the SH-RML. Both FNO and T-UNet models were subjected to 200 steps each for gradient evaluation in the optimization process for each one of the 100 prior models, and the full reservoir simulations with DARTS were also taken into account for the prior and the posterior curves. For SH-RML, it should be noted that the extensive training experiments, which included up to 1000 models, were primarily focused on performance benchmarking. These experiments are not a requirement for the framework but serve to showcase its adaptability and effectiveness in diverse application scenarios. The overall history-matching process requires approximately 8 hours, which is justified by the enhanced accuracy and reliability of the estimates obtained. Nonetheless, the optimization time could potentially be reduced by employing a more tailored approach to the cost function optimization without significantly compromising the quality of the results.

\begin{figure}[ht!]
\centering
\includegraphics[width=1\textwidth]{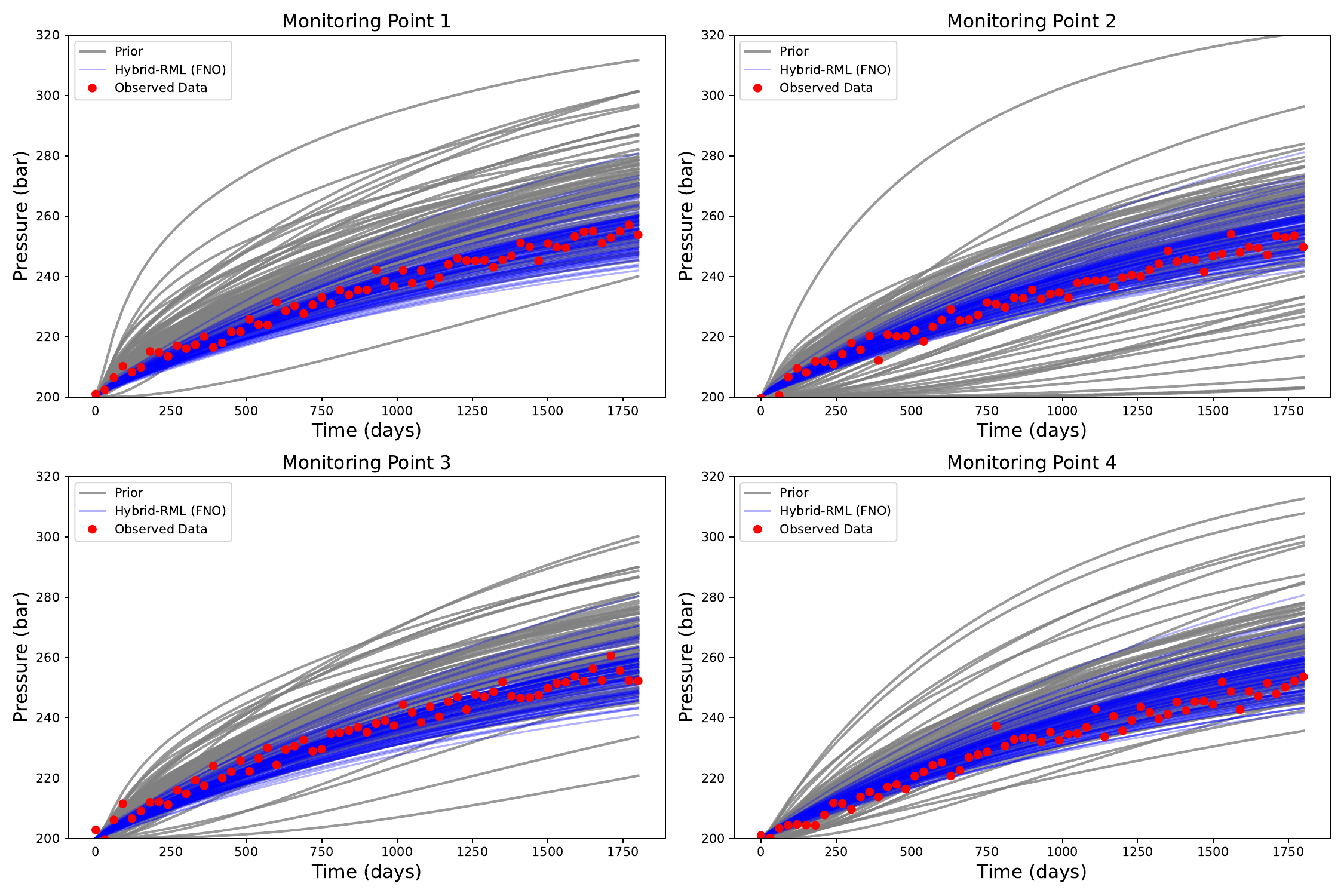}
\caption{Comparison between the prior, reference model, and posterior pressures at each monitoring point for the Hybrid-RML-Surrogate history matching using FNO. Red dots represent a realization of perturbed observed data.}
\label{fig:hrml_FNO_pressure}
\end{figure}

\begin{figure}[ht!]
\centering
\includegraphics[width=1\textwidth]{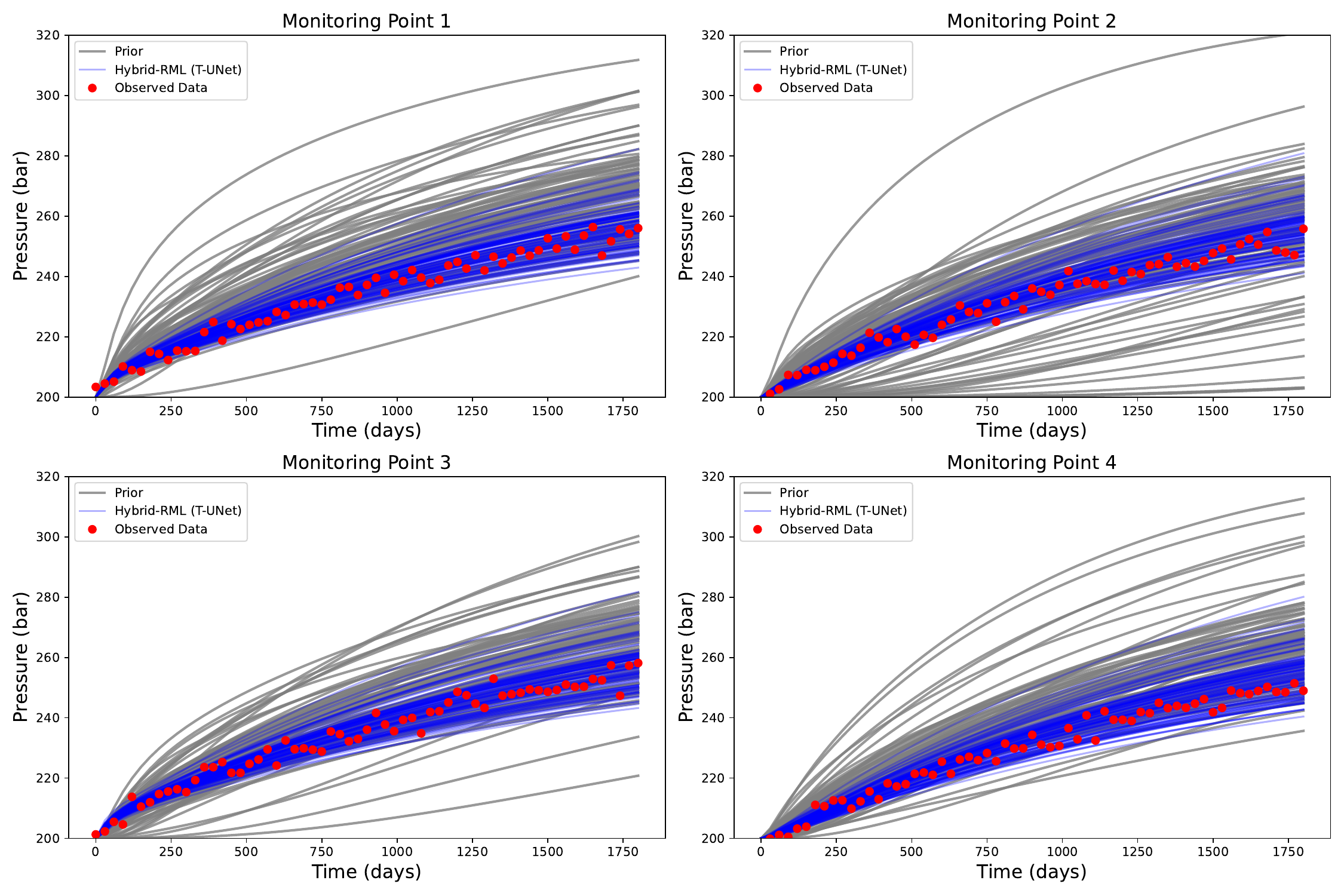}
\caption{Comparison between the prior, reference model, and posterior pressures at each monitoring point for the Hybrid-RML-Surrogate history matching using T-UNet. Red dots represent a realization of perturbed observed data.}
\label{fig:hrml_UNet_pressure}
\end{figure}

\begin{figure}[ht!]
\centering
\includegraphics[width=0.5\textwidth]{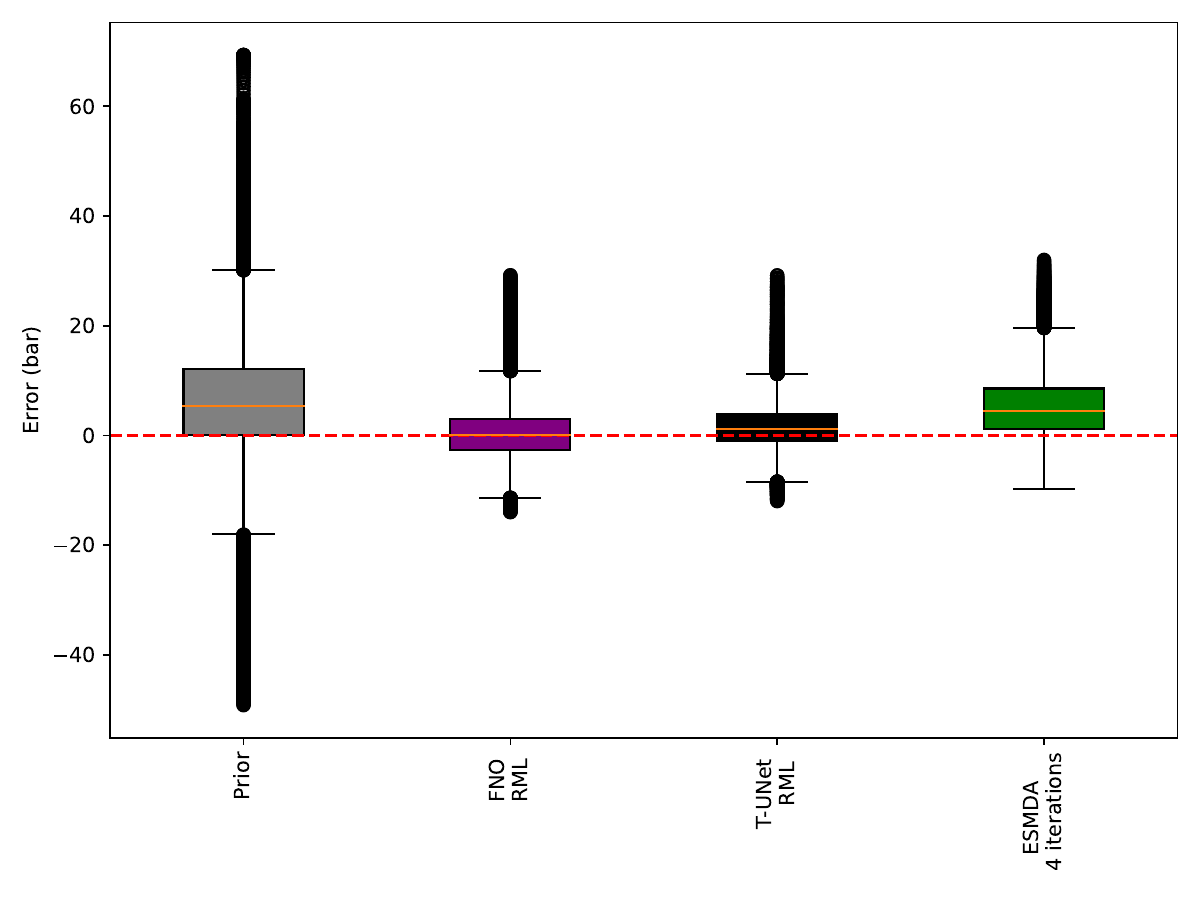}
\caption{Box plots comparing monitoring pressure errors between standard RML and Hybrid-RML-Surrogate using FNO and T-UNet surrogates.}
\label{fig:hrml_error}
\end{figure}

\begin{figure}[ht!]
\centering
\includegraphics[width=0.8\textwidth]{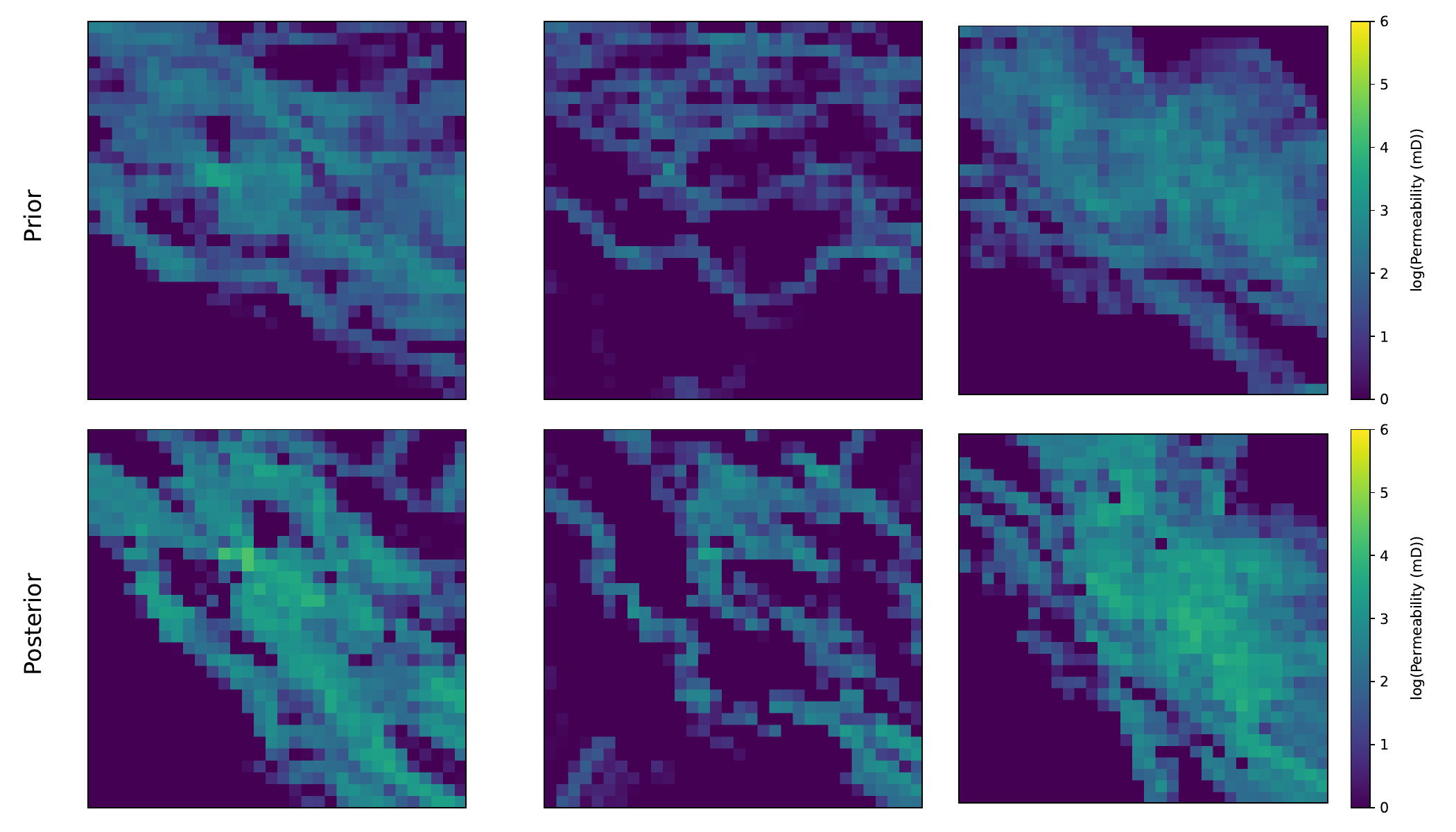}
\caption{Posterior permeabilities for Hybrid-RML-Surrogate with FNO surrogates.}
\label{fig:hrml_perm}
\end{figure}

For SH-RML, we noticed an improvement in history matching with posterior pressure closer to the values of the observations. Figure \ref{fig:hrml_FNO_pressure} and Figure \ref{fig:hrml_UNet_pressure} show the pressure matching results at each monitoring point for SH-RML with FNO, and SH-RML with T-UNet, respectively. While surrogates are employed for gradient evaluations, the final posterior pressure curves are computed using the full reservoir simulator (DARTS). This allows SH-RML to leverage the efficiency of surrogates during optimization while still generating high-fidelity pressure forecasts with the simulator for final uncertainty quantification. Figure~\ref{fig:hrml_error} displays box plots comparing monitoring pressure errors between the standard ESMDA and SH-RML methods. The results from SH-RML exhibit a better balance around zero when compared to ESMDA. Additionally, Figure~\ref{fig:hrml_perm} demonstrates that the posterior permeabilities derived from SH-RML, employing FNO are consistent with the prior. In contrast to ESMDA, RML provides permeabilities within a similar range as the prior. For conciseness and to avoid repetition, we do not include additional figures showcasing the posterior permeabilities using the T-Unet in SH-RML. The outcomes closely mirror those already presented for the FNO version, as well as the enhanced uncertainty quantification compared to ESMDA.

In summary, the results of this case study demonstrate that SH-RML offers improvements over ESMDA for history matching and uncertainty quantification. By leveraging the gradients of ML surrogates for optimization of the cost functions, SH-RML achieves better pressure matching while also providing more accurate estimates of posterior permeability distributions. The integration of surrogates enables the computations of gradient evaluations, enabling thorough optimization of the RML objective.

\section{Discussion} \label{section:discussion}
 
This study presents advancements in combining DA and ML to enhance history matching in CO$_{2}$ storage projects, addressing key challenges and limitations inherent in existing methodologies. Our innovative hybrid frameworks, namely SH-ESMDA and SH-RML, optimize both computational efficiency and accuracy in DA studies. The integration of ML surrogates enabled the application of ensemble-based DA methods like ESMDA efficiently to complex subsurface systems. Specifically, the SH-ESMDA method has accelerated ESMDA computations while maintaining the physical consistency of posterior physical responses. The SH-RML method has excelled in achieving superior history matching compared to standard ESMDA, attributed to the better approximation of the gradients for non-linear and non-Gaussian problems. This method leveraged the automatic differentiation inherent in ML models, enabling gradient-based optimization and overcoming the challenges posed by the absence of adjoint models in reservoir simulators.

A  consideration during this study was the choice of grid resolution. In the preliminary stages, models with resolutions of \(256 \times 256\), \(128 \times 128\), and \(64 \times 64\) were examined. The choice of a \(32 \times 32\) resolution was influenced by extended computational times required for numerous runs at higher resolutions and substantial memory requirements of the FNO. Although subsequent updates to the computational tools have mitigated these challenges, allowing for exploration at higher resolutions, a substantial portion of the work presented was conducted using the initial setup and resolutions due to the constraints at the time. The chosen resolution effectively captures the variability of the permeability of channelized reservoirs, offering a balance between computational demand and model fidelity.

Reflecting on the characteristics and applications of each method, the SH-ESMDA is optimal for scenarios prioritizing computational speed. At the same time, the SH-RML is particularly suited for problems necessitating more accurate gradient computations, even when forward models lack adjoints. The unified parameterization of permeability between ML and physics-based models has facilitated integration, ensuring physical consistency and reliability in the posterior solutions computed using the high-fidelity DARTS simulator.

However, it’s important to acknowledge some limitations of the methods. The SH-ESMDA inherits the fundamental constraints from ESMDA in capturing uncertainties, and the accuracy of both hybrid methods is contingent on the precision of the ML surrogates, which may encounter challenges in extrapolation. Addressing these limitations could potentially be achieved by incorporating more robust and nonlinear DA methods, enhancing the reliability of uncertainty quantification. A limitation of the SH-RML is the dependency on the accuracy of the ML surrogates for gradient computation of specific points of the reservoir. If the surrogates are not well-trained or accurate, the gradients derived from them could be unreliable, leading to suboptimal or incorrect solutions and impacting the method's overall effectiveness.

Our comparative assessment between FNO and T-UNet has demonstrated that both architectures exhibit low RMSE compared to DARTS simulations, with FNO having a slight edge, particularly when training data is limited. This insight is essential for scientists to choose the most suitable ML architectures effectively, especially when considering data availability constraints.
Beyond CO$_{2}$ storage, the versatility of the developed methodologies is evident, with potential adaptability to other applications including geothermal energy and nuclear waste disposal. The advancements made in this study are poised to catalyze the broader adoption and application of hybrid DA-ML methodologies across diverse scientific domains, marking a significant step forward in the field.

\section{Conclusion} \label{section:conclusion}
This paper introduced novel frameworks for enhancing uncertainty quantification (UQ) in Geological Carbon Storage (GCS) projects through the integration of machine learning (ML) and data assimilation (DA) techniques. We evaluated two neural network architectures, the Fourier Meural Operators (FNOs) and Transformer-UNet (T-UNet), for generating accurate and efficient surrogate models of CO\textsubscript{2} injection simulations. Comparative analyses revealed the FNO's slight superiority when training data is limited.

Leveraging these surrogates, we proposed two hybrid methods. Surrogate-based hybrid ESMDA (SH-ESMDA) incorporates the ML models into the Ensemble Smoother with Multiple Data Assimilation (ESMDA), reducing computational time by over 50\% while maintaining accuracy. Surrogate-based Hybrid RML (SH-RML) enables variational data assimilation by using automatic differentiation from the neural networks for gradient calculations in the Randomized Maximum Likelihood (RML) optimization. This avoids manual adjoint derivations.

Results showed that SH-RML achieved improved history matching and uncertainty quantification compared to standard ESMDA. The FNO surrogate enabled efficient computation of gradients for the RML optimization. The proposed frameworks thus enhance DA for GCS by integrating machine learning efficiency with the physical reliability of reservoir simulators.

In conclusion, while our study presents significant advancements in integrating AI with DA for GCS applications, it also opens avenues for future research. Particularly, exploring the scalability of our methodologies to larger, more complex reservoir systems, such as real carbonate formations, and real field cases from current GCS projects and investigating the integration of additional neural network architectures, could yield further valuable insights in optimizing GCS operations. The framework shows versatility beyond CO$_{2}$ sequestration, presenting opportunities for adaptation to other subsurface modeling applications like geothermal energy and nuclear waste storage. Overall, this study demonstrates the potential of combining machine learning and physics-based models to tackle multifaceted problems in uncertainty quantification for the energy transition.

\section*{Acknowledgment}

The authors thank Petróleo Brasileiro S.A. (Petrobras) for sponsoring the doctoral research of Gabriel Serrão Seabra and Vinicius Luiz Santos Silva.

We also thank Guillaume Rongier for providing valuable insights on geological modeling, Alexandre Emerick and Rafael Oliveira for sharing expertise on data assimilation techniques, and Ahmed ElSheikh for fruitful discussions about the machine learning models.

\section*{CRediT authorship contribution statement}

\textbf{Gabriel Serrão Seabra}: Conceptualization, Methodology, Software, Validation, Formal analysis, Investigation, Writing - original draft, Visualization.

\textbf{Nikolaj T. Mücke}: Conceptualization, Methodology, Software, Validation, Writing - review \& editing, Formal analysis, Data curation.

\textbf{Vinicius Luiz Santos Silva}: Conceptualization, Methodology, Validation, Writing - review \& editing.

\textbf{Denis Voskov}: Methodology, Software, Writing - review \& editing, Supervision, Funding acquisition.

\textbf{Femke Vossepoel}: Conceptualization, Methodology, Writing - review \& editing, Supervision, Project administration, Funding acquisition.

\section*{Declaration of competing interest}
The authors declare that they have no competing financial or personal interests that have influenced the work presented in this paper. Gabriel Serrão Seabra and Vinicius Luiz Santos Silva wrote this paper while employed at Petrobras. The company had no role in study design, data collection and analysis, decision to publish, or preparation of the manuscript.

\section*{Declaration of Generative AI and AI-assisted technologies in the writing process}
In the course of preparing this manuscript, the authors utilized ChatGPT from OpenAI to improve readability. Additionally, Grammarly was employed to guarantee grammatical accuracy and to improve overall readability. Subsequent to employing these tools, the authors meticulously reviewed and amended the content as necessary. The authors bear full responsibility for the content of the publication.

\clearpage 
\bibliography{references}

\appendix

\section{Detailed RMSE Metrics for FNO and T-UNet Models}
\label{sec: appendix}
\begin{table}[h!]
\centering
\begin{tabular}{|c|c|c|c|c|}
\hline
Samples & Modes & Width & RMSE (Pressure) & RMSE (CO\textsubscript{2} molar fraction) \\ \hline \hline
\multicolumn{5}{|c|}{FNO} \\ \hline
1000 & 18 & 128 & 4.29081 & 0.01072 \\ \hline
1000 & 18 & 64 & 4.41009 & 0.01047 \\ \hline
1000 & 6 & 128 & 4.28229 & 0.00895 \\ \hline
1000 & 6 & 64 & 4.66067 & 0.00759 \\ \hline
500 & 18 & 128 & 5.80695 & 0.01089 \\ \hline
500 & 18 & 64 & 5.44274 & 0.01092 \\ \hline
500 & 6 & 128 & 6.76917 & 0.01074 \\ \hline
500 & 6 & 64 & 6.54552 & 0.01087 \\ \hline
200 & 18 & 128 & 8.22966 & 0.01156 \\ \hline
200 & 18 & 64 & 8.34038 & 0.01713 \\ \hline
200 & 6 & 128 & 8.29742 & 0.01250 \\ \hline
200 & 6 & 64 & 8.20080 & 0.01262 \\ \hline
100 & 18 & 128 & 9.91828 & 0.01761 \\ \hline
100 & 18 & 64 & 10.07389 & 0.01789 \\ \hline
100 & 6 & 128 & 8.75043 & 0.01727 \\ \hline
100 & 6 & 64 & 10.90478 & 0.01747 \\ \hline
\multicolumn{5}{|c|}{T-UNet} \\ \hline
1000 & - & - & 2.27017 & 0.00620 \\ \hline
500 & - & - & 6.04151 & 0.00646 \\ \hline
200 & - & - & 12.78398 & 0.01721 \\ \hline
100 & - & - & 12.71158 & 0.03029 \\ \hline

\end{tabular}
\caption{Test RMSE Metrics for Pressure and CO\textsubscript{2} molar fraction}
\label{tab:unified_model_metrics}
\end{table}

\end{document}